\documentclass[final]{cvpr}

\usepackage{times}
\usepackage{epsfig}
\usepackage{graphicx}
\usepackage{amsmath}
\usepackage{amssymb}

\usepackage{subfigure}
\usepackage{multicol,multicap}
\usepackage{algorithm}
\usepackage{algorithmic}
\usepackage{booktabs,tabularx,multirow}

\usepackage[pagebackref=true,breaklinks=true,colorlinks,bookmarks=false]{hyperref}

\def\cvprPaperID{368} 
\def\confYear{CVPR 2021}

\begin{document}

\title{Dense Label Encoding for Boundary Discontinuity Free Rotation Detection}

\author{Xue Yang$^{1}$, Liping Hou$^{2}$, Yue Zhou$^{1}$, Wentao Wang$^{1}$, Junchi Yan$^{1,}$\thanks{Corresponding author is Junchi Yan.}\\
$^{1}$Shanghai Jiao Tong University, Shanghai, China\\
$^{2}$University of Chinese Academy of Sciences, Beijing, China\\
{\tt\small \{yangxue-2019-sjtu, sjtu\_zy, wwt117, yanjunchi\}@sjtu.edu.cn} \\
{\tt\small houliping17@mails.ucas.ac.cn}
}

\maketitle

\begin{abstract}
   Rotation detection serves as a fundamental building block in many visual applications involving aerial image, scene text, and face etc. Differing from the dominant regression-based approaches for orientation estimation, this paper explores a relatively less-studied methodology based on classification. The hope is to inherently dismiss the boundary discontinuity issue as encountered by the regression-based detectors. We propose new techniques to push its frontier in two aspects: i) new encoding mechanism: the design of two Densely Coded Labels (DCL) for angle classification, to replace the Sparsely Coded Label (SCL) in existing classification-based detectors, leading to three times training speed increase as empirically observed across benchmarks, further with notable improvement in detection accuracy; ii) loss re-weighting: we propose Angle Distance and Aspect Ratio Sensitive Weighting (ADARSW), which improves the detection accuracy especially for square-like objects, by making DCL-based detectors sensitive to angular distance and object's aspect ratio. Extensive experiments and visual analysis on large-scale public datasets for aerial images i.e. DOTA, UCAS-AOD, HRSC2016, as well as scene text dataset ICDAR2015 and MLT, show the effectiveness of our approach. The source code is available at \url{https://github.com/Thinklab-SJTU/DCL_RetinaNet_Tensorflow} and is also integrated in our open source rotation detection benchmark: \url{https://github.com/yangxue0827/RotationDetection}.
\end{abstract}

\section{Introduction}
Rotation detection has recently attracted increasing attention for their important utility across different scenarios, including aerial images, scene text, and faces etc., which is relatively less studied compared with the vast literature in horizental object detectors that do not estimate the exact rotation while only output the horizontal bounding box.


Many mainstream rotation detection algorithms (including aerial image \cite{yang2018automatic, jiao2018densely, yang2018position, azimi2018towards, ding2018learning, yang2019scrdet, yang2019r3det, qian2019learning, yang2020arbitrary, yang2020scrdet++, yang2020on}, scene text \cite{zhou2017east, liu2018fots, jiang2017r2cnn, ma2018arbitrary, liao2018rotation, liao2018textboxes++} and face \cite{shi2018real, huang2007high, rowley1998rotation}) are derived based on the vanilla detection algorithms \cite{girshick2015fast,ren2015faster, redmon2016you,dai2016r,lin2017feature,lin2017focal}. Among them, the rotation detection algorithm based on five parameters ($[x,y,h,w,\theta]$) dominates. Similar to the coordinate regression method in horizontal detection, angle parameter is also predicted by regression. Although gratifying results have been achieved, there are still some fundamental flaws in the orientation estimation based on regression. Angle prediction based on regression often introduces boundary discontinuity \cite{yang2019scrdet, qian2019learning, yang2020arbitrary, yang2020on}, mainly including periodicity of angle (PoA) and exchangeability of edges (EoE). The main reason for the former is the bounded periodic nature of the angle parameter, while the latter is related to the definition of the bounding box. In general, the root cause is that the ideal predictions are beyond the defined range. Due to the sharp increase in the loss at the boundary, the regression form of the model at the boundary and non-boundary can not be consistent. Therefore, the model has to predict the angle in a more complicated form at the boundary, which increases the burden of the model and also increases the difficulty of prediction at the boundary. This is fatal for the rotation object detection that require high precision, especially for objects with large aspect ratios. 

Most existing works aim to eliminate the sudden loss increase by adding constraints on the loss function or changing the way of calculation, such as IoU Smooth L1 Loss \cite{yang2019scrdet} and Modulated Loss \cite{qian2019learning}, as shown in Table \ref{table:peer_methods}. The advantage is that they can borrow the well developed baselines from horizontal object detectors and reuse the related techniques to boost the detection performance. However, such ad-hoc techniques applied on the regression-based detectors cannot guarantee the full dismiss of boundary discontinuity behavior. To give a more elegant and effective solution, the more recent work called Circular Smooth Label (CSL) \cite{yang2020arbitrary, yang2020on} argues to apply angle classification instead of regression to address PoA. Then, CSL-based method combines with the long-side definition (five-parameter with $180^\circ$ angular range) of bounding box to further tackle the EoE problem. The use of `CSL+$180^\circ$' \footnote{Unless otherwise specified, the CSL-based method mentioned in this paper is based on the long-side definition method.} leads to a natural solution to get rid of the boundary discontinuity issue. As angle classification based detectors are still in its early stage, there are still many limitations e.g. very heavy prediction layer and difficulty in handling square-like objects. The former problem is initially explored by the study \cite{yang2020on}. This paper is one of the classification-based endeavors in pushing forward this frontier, with two specific technical advancements as follows.

\begin{table}[!tb]
	\centering
	\resizebox{0.48\textwidth}{!}{
		\begin{tabular}{cccccccc}
			\toprule
			Method & Box Def. & Angle Pred. & PoA & EoE & SLP & Speed & mAP$_{50:95}$ \\
            \hline
            RetinaNet & Long-Side Def. & Reg. & $\checkmark$ & $\checkmark$ & $\checkmark$ & - & 31.49\\
            RetinaNet & OpenCV Def. & Reg. & $\checkmark$ & $\checkmark$ & $\times$ & $\sim$1x & 34.50\\
            IoU-Smooth L1 Loss \cite{yang2019scrdet} & OpenCV Def. &  Reg. & $\times$ & $\times$ & $\times$ & $\sim$1x & 36.23\\ 
            Modulated Loss \cite{qian2019learning} & OpenCV Def. & Reg. & $\times$ & $\times$ & $\times$ & $\sim$1x & 34.61\\
            CSL \cite{yang2020arbitrary} & Long-Side Def. & Cls.: SCL & $\times$ & $\times$ & $\checkmark$ & $\sim \frac{1}{3}$x & 35.04\\ 
            DCL (BCL) & Long-Side Def. & Cls.: DCL & $\times$ & $\times$ & $\times$ & $\sim$1x & \textbf{36.71}\\ 
			\bottomrule
	\end{tabular}}
		\vspace{-10pt}
	\caption{Comparison between different solutions for periodicity of angle (PoA), exchangeability of edges (EoE) and square-like problem (SLP) on DOTA val set. The $\checkmark$ indicates that the method suffers the corresponding problem.}
	\label{table:peer_methods}

\end{table}

\begin{figure}[!tb]
		\centering
		\subfigure[RetinaNet-Reg]{
			\begin{minipage}[t]{0.46\linewidth}
				\centering
				\includegraphics[width=0.98\linewidth]{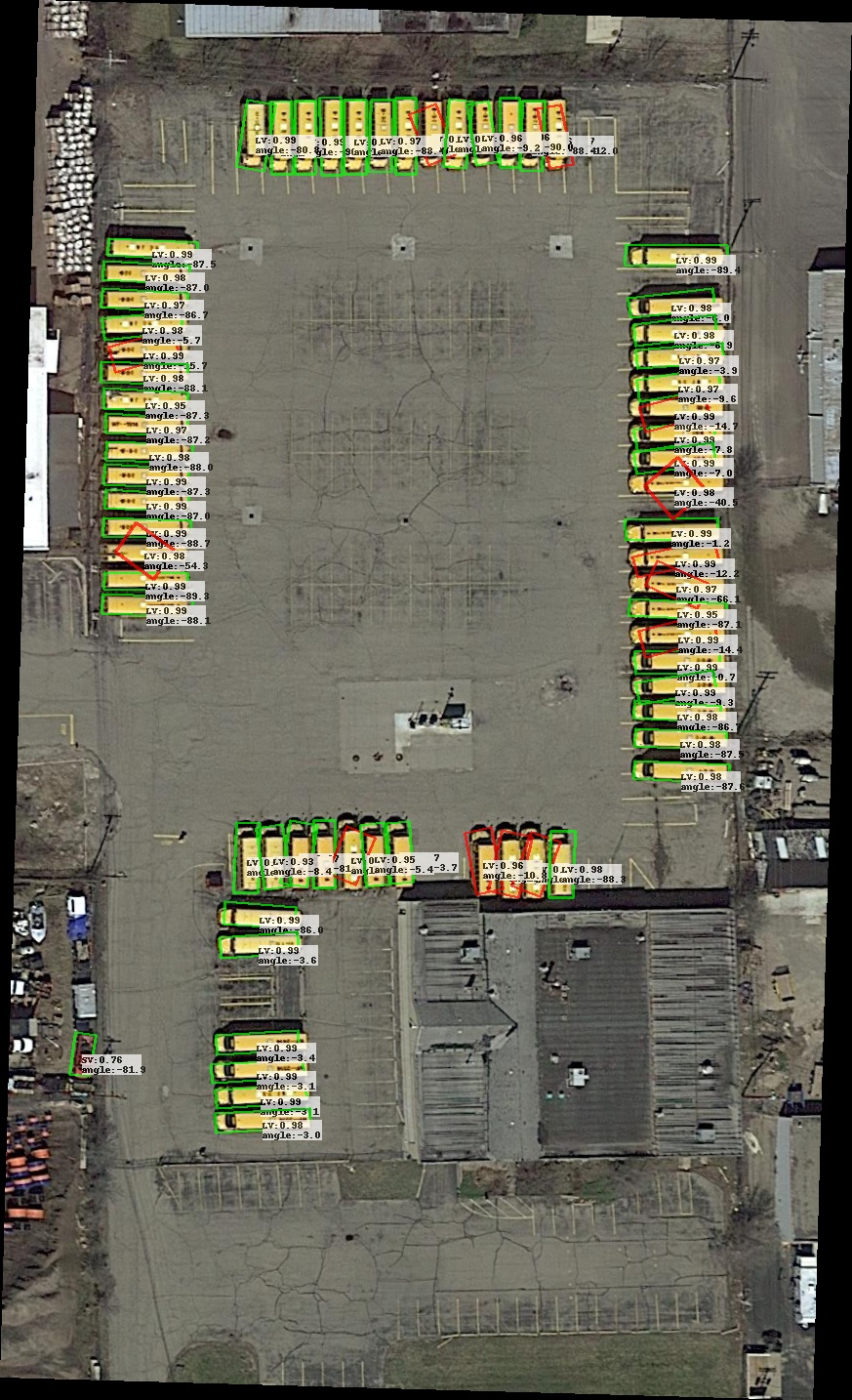}
			\end{minipage}%
			\label{fig:P0016_reg}
		}
		\subfigure[RetinaNet-CSL]{
			\begin{minipage}[t]{0.46\linewidth}
				\centering
				\includegraphics[width=0.98\linewidth]{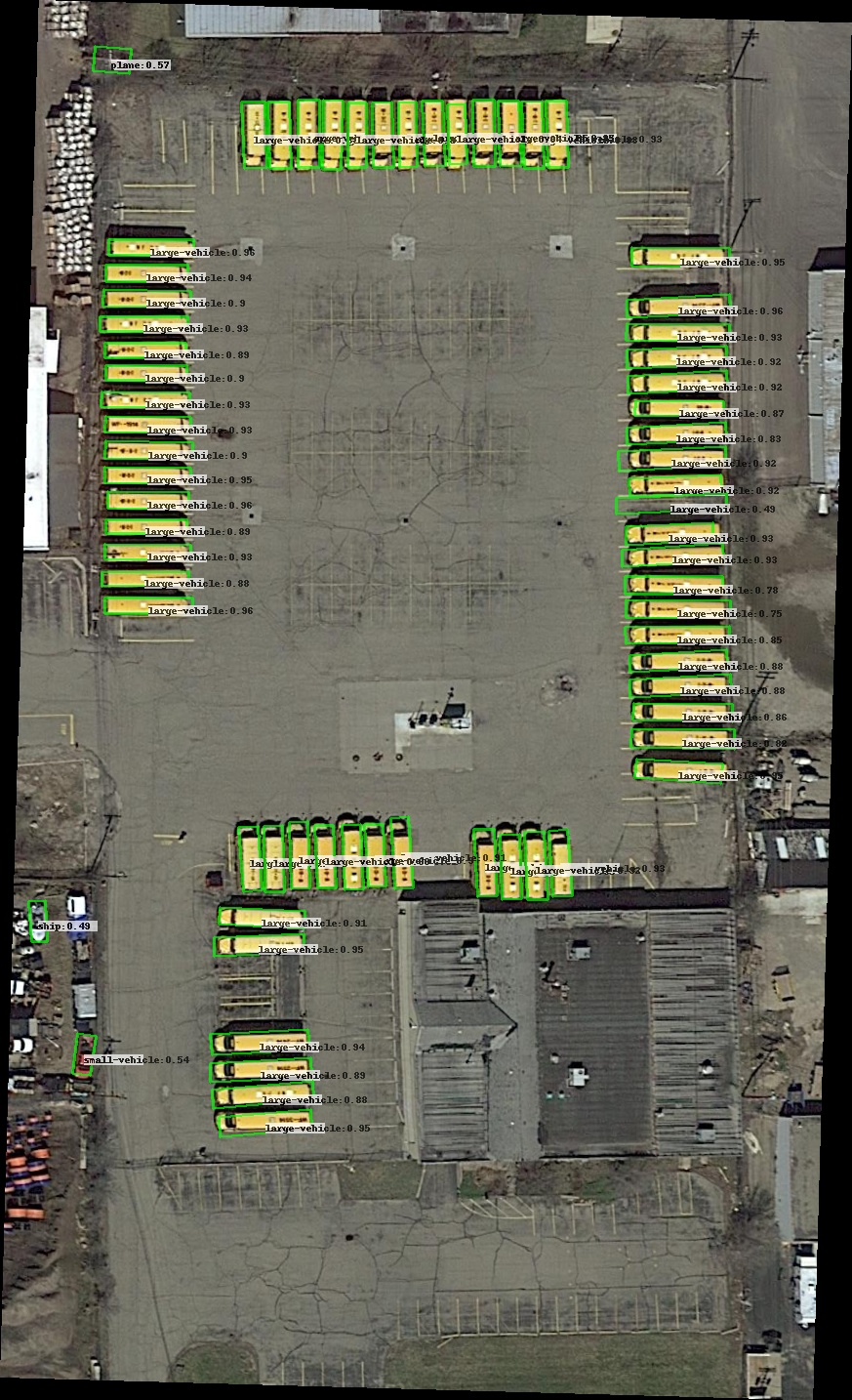}
			\end{minipage}%
			\label{fig:P0016_csl}
		}\\
		\subfigure[RetinaNet-BCL]{
			\begin{minipage}[t]{0.46\linewidth}
				\centering
				\includegraphics[width=0.98\linewidth]{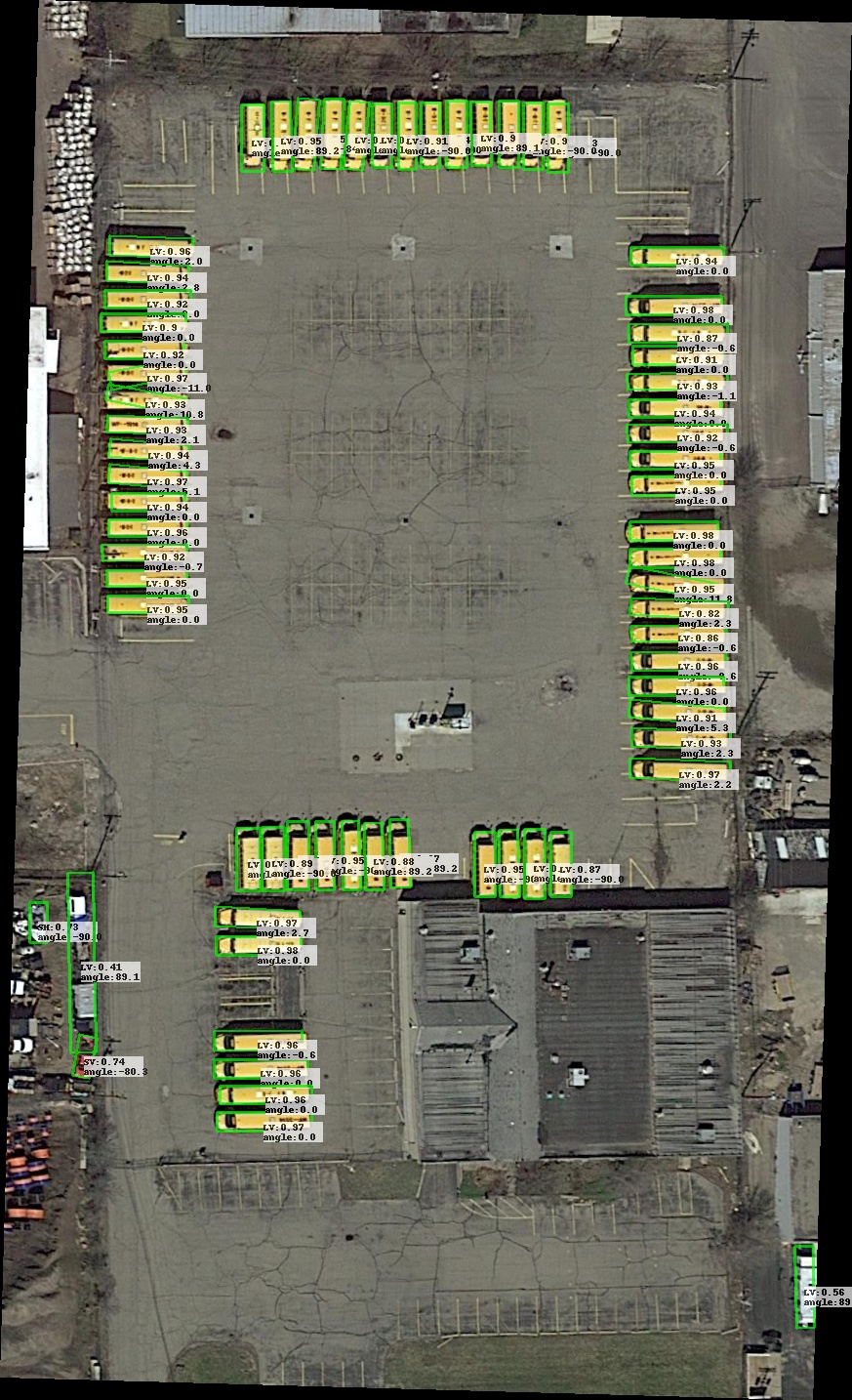}
			\end{minipage}
			\label{fig:P0016_bcl}
		}
		\subfigure[RetinaNet-GCL]{
			\begin{minipage}[t]{0.46\linewidth}
				\centering
				\includegraphics[width=0.98\linewidth]{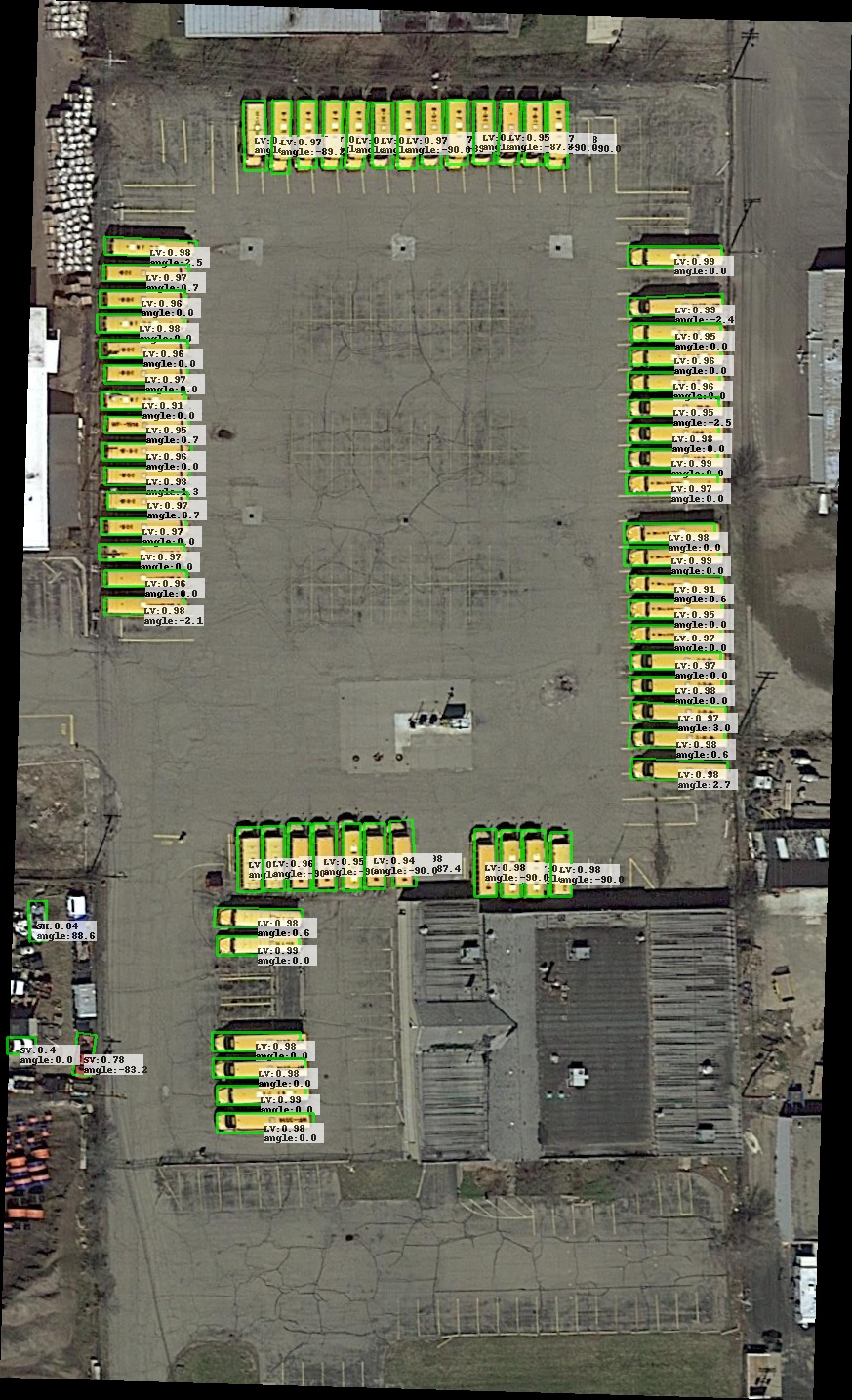}
			\end{minipage}%
			\label{fig:P0016_gcl}
		}
		\centering
		\caption{Comparison of four rotation detectors in the boundary case. Red bounding boxes indicate some bad detection cases (zoom in for better view). CSL and DCL based methods (including BCL and GCL) are totally boundary discontinuity free.}
		\label{fig:bc_vis}
		\vspace{-8pt}
\end{figure}


First, we design two Densely Coded Labels (DCL) in contrast to the Sparsely Coded Label (SCL, including CSL, One-Hot encoding), which has empirically led to notable training time reduction with meanwhile improved detection accuracy. To make DCL as sensitive to angle distance as CSL, we calculate the decimal difference between the predicted angle and the angle label as a angle distance aware weight. However, this weight will reintroduce the PoA problem, and we find that the long-side definition method is not conducive to the training of square-like objects. Based on the findings of the above two issues, we design Angle Distance and Aspect Ratio Sensitive Weighting (ADARSW). ADARSW can eliminate the PoA and can be adaptively adjusted according to the object's aspect ratio, which can greatly reduce the burden of model training. Combing `DCL+$180^\circ$+ADARSW' as a whole, extensive experiments and visual analysis on different datasets and detectors prove that DCL-based method can be a better baseline choice than the angle regression-based and CSL-based methods. In summary, our work makes the following contributions:

1) To improve the robustness especially for objects with small aspect ratio, we propose Angle Distance and Aspect Ratio Sensitive Weighting (ADARSW), which further improves accuracy by making our proposed DCL-based detector sensitive to angular distance and object's aspect ratio. In contrast, the existing CSL-based detector suffers from its long-side definition for detecting square-like objects.

2) We compare the impact of two classic Densely Coded Labels (DCL) by introducing them to the angle classification task for potential speedup, namely Binary Coded Label (BCL) and Gray Coded Label (GCL), which are more compact than existing CSL. Though GCL has been partly studied in \cite{yang2020on}, while this paper presents a more thorough investigation especially for BCL. We empirically show that DCL, especially BCL can lead to notable training speed boost (about three times) as well as detection accuracy.


3) Extensive experiments and visual analysis on different datasets and detectors prove the efficacy of our techniques. It outperforms state-of-the-art CSL-based detector~\cite{yang2020arbitrary} by: 77.37\% vs. 76.17\% on DOTA dataset.

\section{Related Work}
\paragraph{Horizental Object Detection}
Object detection is a fundamental task in the field of computer vision, and it has developed rapidly in recent years. Classic convolutional neural networks (CNN) based detectors can mainly be divided into two categories: two-stage object detectors \cite{girshick2015fast,ren2015faster,dai2016r,lin2017feature} and single-stage object detectors \cite{redmon2016you,liu2016ssd,lin2017focal}. Two-stage methods are based on region proposal and have achieved promising results on some benchmarks, whereas single-stage approaches simplify detection as a regression problem to maintain a faster speed. Compared to anchor-based methods, many anchor-free based methods have become extremely popular in recent years. CornerNet \cite{law2018cornernet}, FCOS \cite{tian2019fcos}, CenterNet \cite{duan2019centernet} and ExtremeNet \cite{zhou2019bottom} attempt to predict some keypoints of objects such as corners or extreme points, which are then grouped into bounding boxes. What is even more surprising is that DETR \cite{carion2020end} has constructed a new object detection paradigm based on transformer \cite{vaswani2017attention}, which achieves anchor free and non maximum suppression (NMS) free at the same time. However, horizental object detectors cannot provide accurate orientation and scale information, so they cannot be directly applied to some specific scenes, such as aerial images and scene text.

\paragraph{Rotation Object Detection}
Rotation detectors are mainly applied in the aerial images and scene text. Recent advances in multi-oriented object detection are mainly driven by adaption of classical object detectors using rotated bounding boxes or quadrangles to represent multi-oriented objects. In the aerial imagery scene, ICN \cite{azimi2018towards}, ROI-Transformer \cite{ding2018learning}, CAD-Net\cite{zhang2019cad}, SCRDet \cite{yang2019scrdet}, R$^3$Det \cite{yang2019r3det}, and CSL \cite{yang2020arbitrary} achieve promising performance. Gliding Vertex \cite{xu2020gliding} and RSDet \cite{qian2019learning} achieve more accurate object detection through quadrilateral regression prediction. RRPN \cite{ma2018arbitrary}, TextBoxes++ \cite{liao2018textboxes++} and RRD \cite{liao2018rotation} and FOTS \cite{liu2018fots} are some advanced methods for scene text detection. However, most of the above regression-based arbitrary-oriented methods focus on the prediction of angle using regression yet ignore the boundary discontinuity. Although SCRDet and RSDet solve the boundary discontinuity from the perspective of the loss function, they are not truly boundary discontinuity free methods. CSL-based detector is a new boundary discontinuity free rotation detector, which transforms angular prediction from a regression problem to a classification problem. However, CSL-based method needs to face two obvious shortcomings: heavy prediction layer and unfriendly to square-like objects. In this paper, we aim to solve the above problems from the two perspectives of encoding form and loss function weight.

\begin{figure}[!tb]
	\centering
	\includegraphics[width=0.9\linewidth]{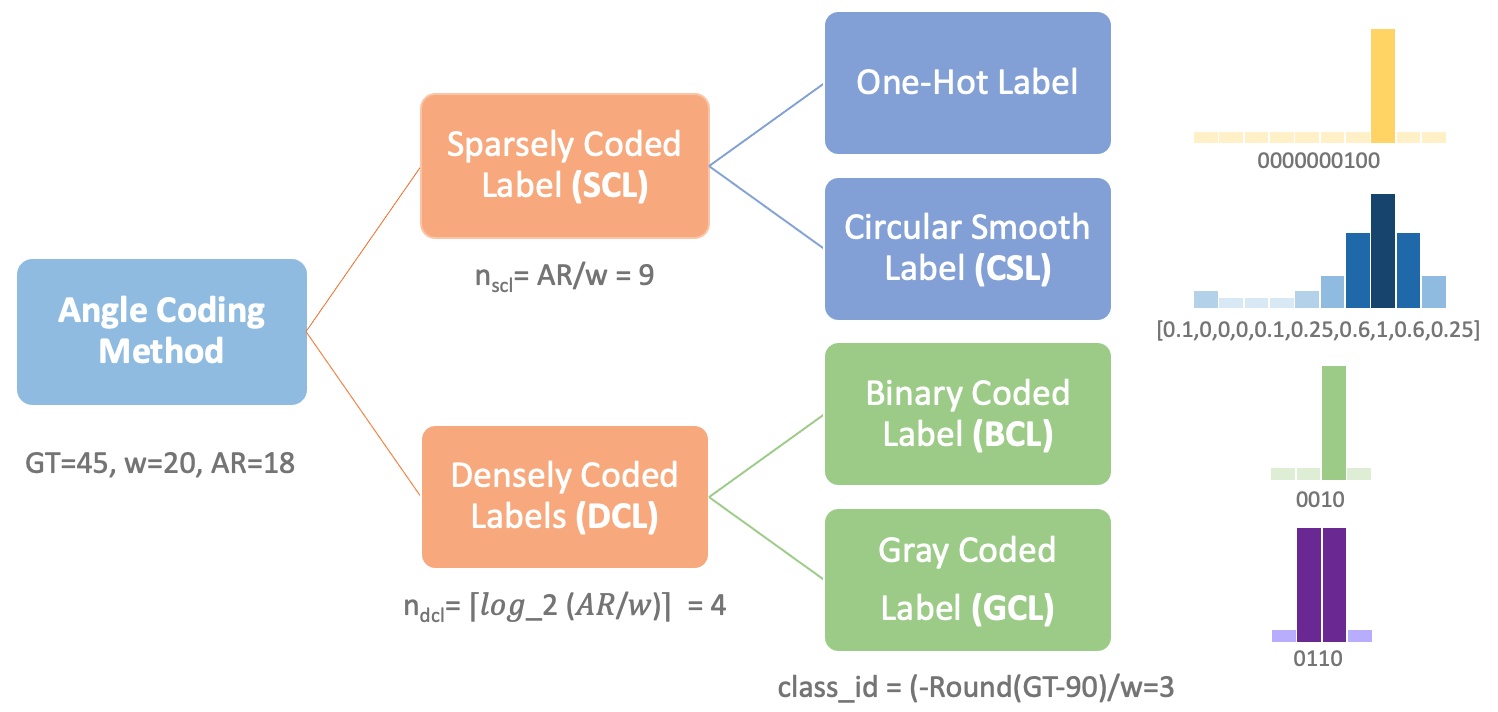}
	\caption{The relationship between the various angle encoding methods.}
	\label{fig:coding_relation}
	\vspace{-10pt}
\end{figure}

\section{Mitigating Boundary Discontinuity by Classification}\label{sec:bkgn}
The boundary discontinuity \cite{yang2019scrdet, yang2020arbitrary, yang2020on} usually refers to the sharp loss increase of the regression-based rotation detector at the boundary situation, which makes the model unable to perform regression prediction in the same ideal and simple form at the boundary as at the non-boundary. The reasons for the boundary discontinuity are related to the definition of the object bounding box. The work \cite{yang2020arbitrary} summarizes several commonly used bounding box definitions and the causes of boundary discontinuity corresponding to these methods. Details are as follows:

1) Five-parameter method with $90^\circ$ angular range (OpenCV definition method): mainly including periodicity of angular (PoA) and exchangeability of edges (EoE).

2) Five-parameter method with $180^\circ$ angular range (long-side definition method): mainly suffer from periodicity of angular.

3) Eight-parameter method: sorting of the four corners.

The main cause of boundary discontinuity based on regression methods is that the ideal predictions are beyond the defined range. As a consequence, the detection result at the boundary of the detector that has not solved the boundary discontinuity often shows inaccurate angle prediction, as shown in the red bounding boxes in Figure \ref{fig:P0016_reg}. Different solutions are proposed, such as constraining the loss function \cite{qian2019learning, yang2019scrdet}, changing the angle prediction form \cite{yang2020arbitrary,yang2020on}, etc. Although some progress has been made, these methods have their own weaknesses. Many current methods have not notice or eliminate boundary discontinuity from method design. A true boundary discontinuity free detector will provide a more robust high-performance baseline, so designing such detector is a valuable research direction.

\begin{figure}[!tb]
	\centering
		\begin{multicols}{2}
		
		\subfigure[DCL: Binary Coded Label]{
			\includegraphics[width=0.90\linewidth]{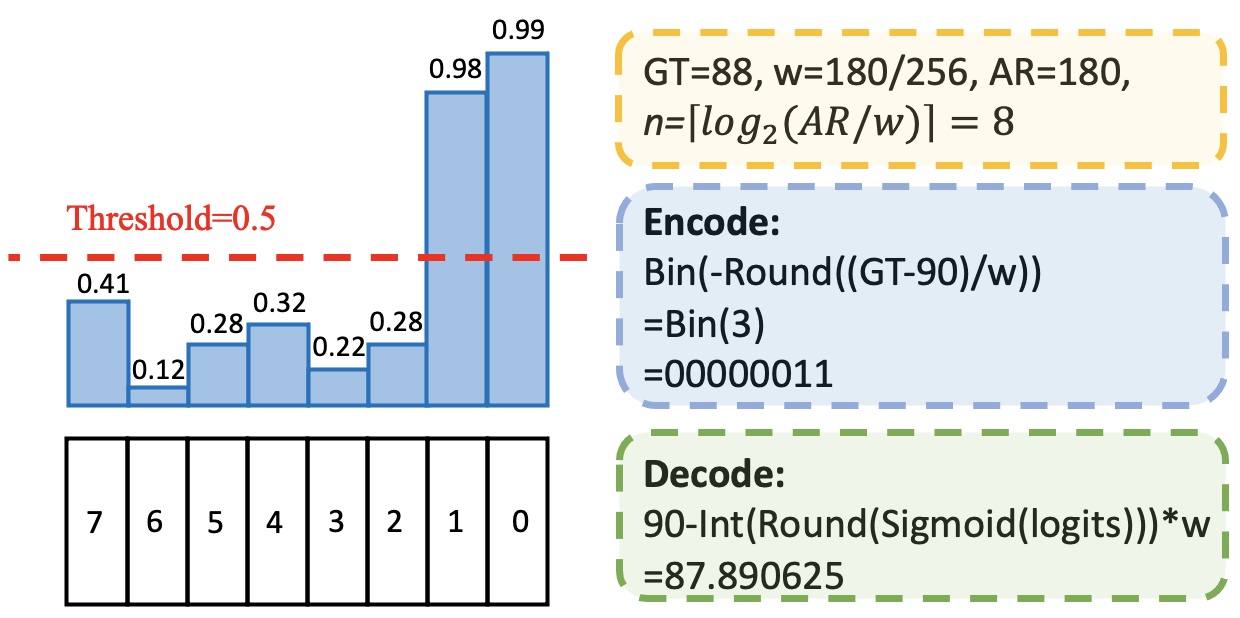}
			\label{fig:dcl}

		}
		
        \subfigure[SCL: One-Hot Label]{
			\includegraphics[width=0.90\linewidth]{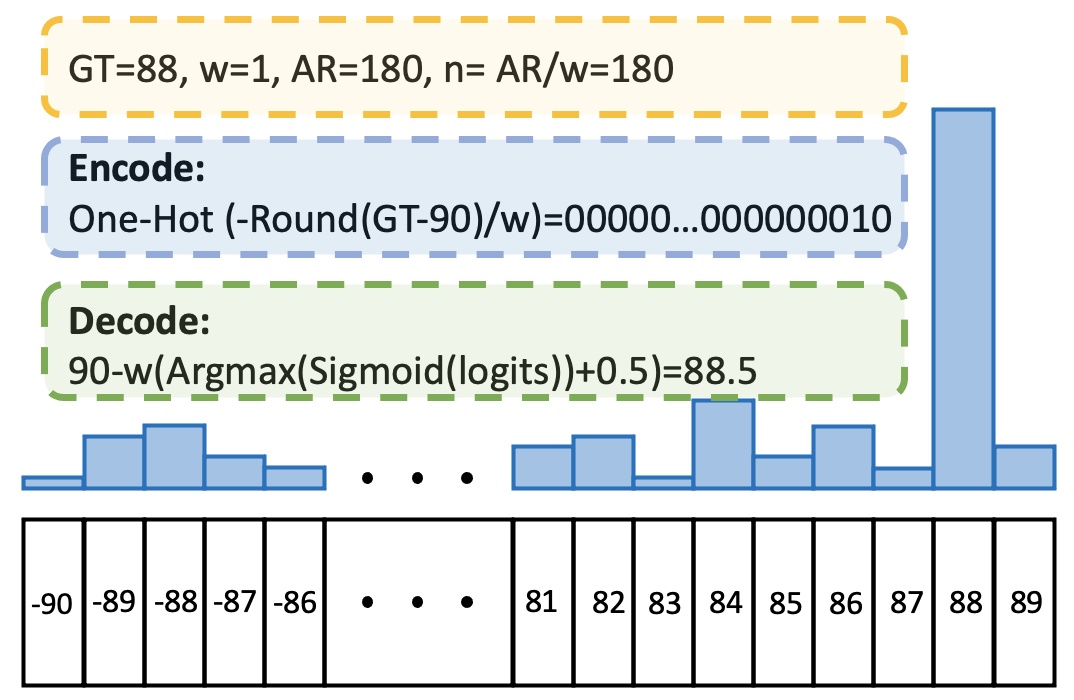}
			\label{fig:onehot}

		}
        \subfigure[SCL: Circular Smooth Label]{
			\includegraphics[width=0.92\linewidth]{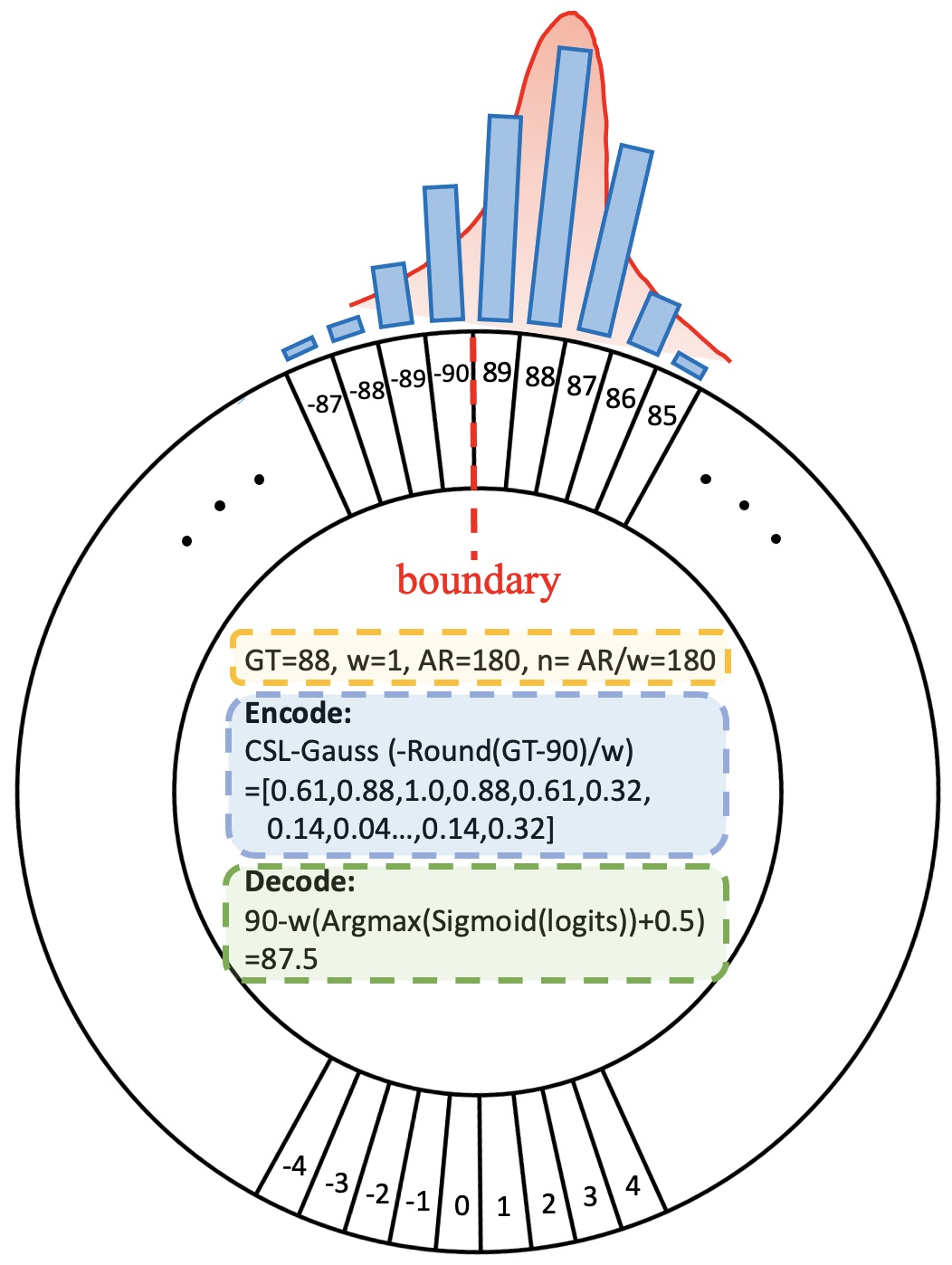}
			\label{fig:csl}

		}
        \end{multicols}
	\centering
	\caption{Examples  of  encoding  and  decoding  process  of  One-Hot, CSL-Gaussian and BCL for angle prediction.}
	\label{fig:angle_label}
	\vspace{-10pt}
\end{figure}

\section{Proposed Method}
In this section, we first give a retrospection to the recent classification-based rotation detectors namely  Circular Smooth Label (CSL) \cite{yang2020arbitrary}, pointing out its limitation in achieving a cost-efficient detector. Then we propose our so-called Densely Coded Label (DCL) technique to improve the efficiency and also develop the Angle Distance and Aspect Ratio Sensitive Weighting technique, to improve its sensitivity to small aspect ratio objects. Figure \ref{fig:coding_relation} shows the relationship between various angle encoding methods.

\subsection{Rethinking on Sparsely Coded Label Encoding}\label{sec:scl}
Instead of using the regression-based loss function, the Circular Smooth Label (CSL) \cite{yang2020arbitrary} detectors have been recently proposed which transform rotation detection to a classification task such that the boundary issue naturally disappear. The CSL-based detectors adopt the so-called Sparsely Coded Label (SCL) encoding technique~\cite{yang2020arbitrary} to discretize the angle into a finite number of intervals, and then predicts a discrete angle by classification\footnote{CSL can only solve the PoA, and the EoE problem can be solved by the $180^\circ$ angular definition method (recall the discussion in Section~\ref{sec:bkg}).}. Equation \ref{eq:encode_decode_csl} describes the angle prediction process in CSL:
\begin{equation}
	\begin{aligned}
	 \textbf{Encode:} & \ \text{CSL}( -\text{Round}((\theta_{gt}-90)/\omega)) \\
	 \textbf{Decode:} & \ 90 - \omega (\text{Argmax}(\text{Sigmoid}(logits)) + 0.5)
	\label{eq:encode_decode_csl}
	\end{aligned}
\end{equation}
where $\theta_{gt}$ presents the angle decimal label, $\omega=AR/C_{\theta}$ indicates the angle discretization granularity. $AR$ and $C_{\theta}$ represents angle range (the default value is 180) and the number of angle categories, respectively.  

Figure \ref{fig:onehot} and Figure \ref{fig:csl} show examples of encoding and decoding process of One-Hot and CSL-Gaussian for angle prediction, both of which are embodiments of the SCL encoding. Although the combined techniques `CSL+$180^\circ$' can well eliminate the impact of boundary discontinuity, it also brings two thorny problems that hurt the efficiency and efficacy: i) {a very heavy prediction layer} and ii) {unfriendliness to objects with small aspect ratio}.

\textbf{Thick prediction layer.} 
Equation \ref{eq:thickness} compares the prediction layer thickness of three angle prediction methods:
\begin{equation}
	\begin{aligned}
	\text{Th}_{reg.} = & A \\
	\text{Th}_{onehot}= & \text{Th}_{csl}= A \times AR/\omega
	\label{eq:thickness}
	\end{aligned}
\end{equation}
where $A$ indicates the number of anchors.

\begin{figure}[!tb]
		\centering
		\subfigure[Ground Truth]{
			\begin{minipage}[t]{0.47\linewidth}
				\centering
				\includegraphics[width=0.98\linewidth]{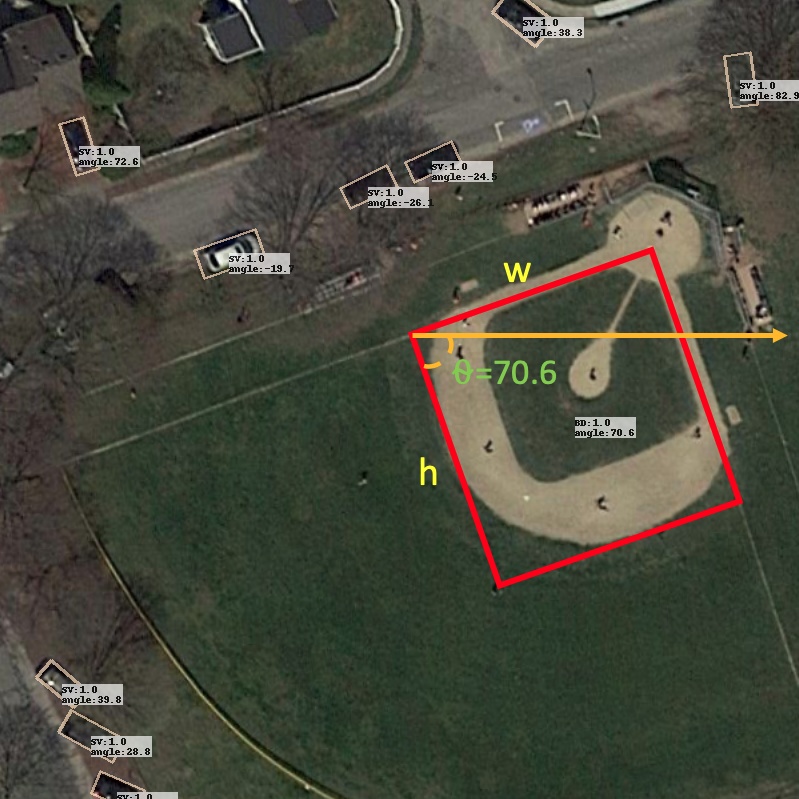}
			\end{minipage}%
			\label{fig:gt_180_BD}
		}
		\subfigure[Prediction after using ADARSW]{
			\begin{minipage}[t]{0.47\linewidth}
				\centering
				\includegraphics[width=0.98\linewidth]{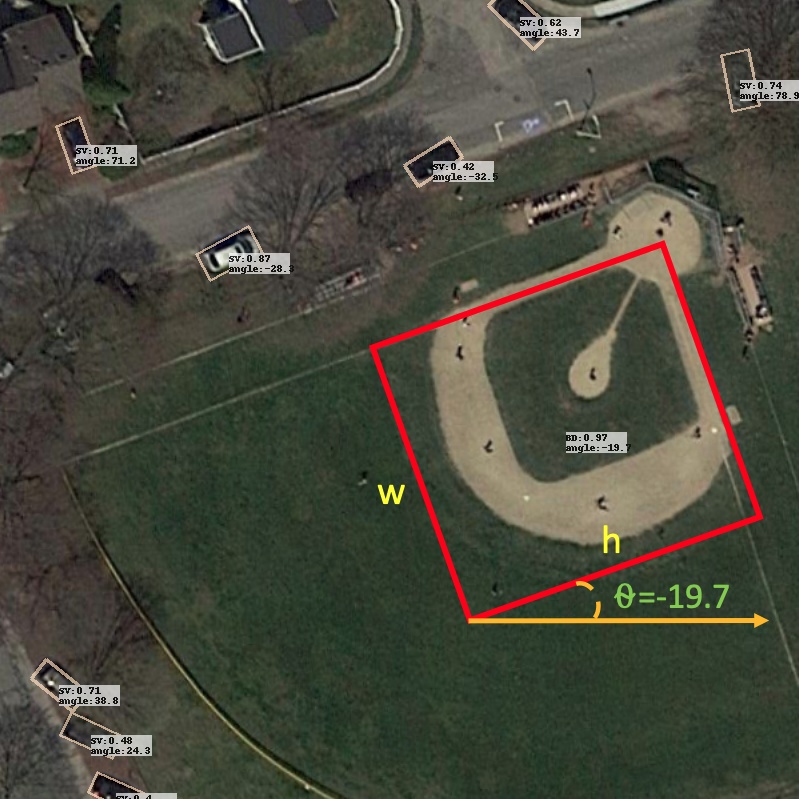}
			\end{minipage}%
			\label{fig:pred_180_BD}
		}
		\centering
		\caption{Illustration for the limitation of the long-edge definition method on the square-like objects. High IoU, but large training loss due to angle difference.}
		\label{fig:180_BD}
\end{figure}
Taking $A=21$, $AR=180$, $w=1$ as an example, the thickness of the prediction layer required by CSL and One-Hot is $3,780$, while the thickness of regression-based approach is only $21$. From the perspective of GFlops and Param, detectors based on CSL have increased by about 82.96\% and 23.42\% respectively. In addition, the training time of RetinaNet-CSL is three times longer than regression-based detector. The detailed statistical results are shown in Table \ref{table:flops_param}. 

\begin{table}[tb!]
	\centering
	\resizebox{0.48\textwidth}{!}{
		\begin{tabular}{cccrcrc}
			\toprule
			Base Model & $\omega$ & GFlops & $\Delta$GFlops & Params (M) & $\Delta$Params & Training Time \\
			\midrule
			
			RetinaNet-Reg & - & 139.35 & - & 36.97 & - & - \\
			RetinaNet-CSL & 1 & 254.96 & +82.96\% & 45.63 & +23.42\% & $\sim$3x \\
			RetinaNet-BCL & 1 & 143.87 & \textbf{+3.24\%} & 37.31 & \textbf{+0.92\%} & \textbf{$\sim$1x} \\
			RetinaNet-GCL & 1 & 143.87 & \textbf{+3.24\%} & 37.31 & \textbf{+0.92\%} & \textbf{$\sim$1x} \\
			\bottomrule
			
	\end{tabular}}
		\vspace{-6pt}
	\caption{Comparison of GFlops and Param over rotation detectors, under the same setting and hyperparameters.}
	\label{table:flops_param}
	\vspace{-10pt}
\end{table}

\textbf{Unfriendliness to small aspect ratio objects.} 
Five-parameter method with $180^\circ$ angular range is a widely used rectangular definition (long-side definition method, $[x,y,h,w,\theta]$) without EoE problem. The $\theta$ is determined by the long side ($h$) of the rectangle and x-axis. However, this definition method is not suitable for square-like box and will suffer a special problem, as shown in Figure \ref{fig:180_BD}. Figure \ref{fig:gt_180_BD}-\ref{fig:pred_180_BD} are ground truth and candidate prediction bounding box with an aspect ratio close to 1, and their angles are $70.6^\circ$ and $-19.7^\circ$, respectively. By calculating the Intersection-over-Union (IoU) and regression (e.g. smooth l1) or classification (e.g. CSL) loss of these two boxes, we find that the IoU between them is close to 1, but a relatively large loss value is produced. This loss value mainly comes from the angle parameter. Therefore, the prediction results in Figure \ref{fig:pred_180_BD} are not allowed by the model, which is too harsh and increases the model's difficulty in predicting objects with small aspect ratio. In fact, this phenomenon becomes less noticeable as the aspect ratio increases. For the definition of square-like objects, using the OpenCV definition method with a period of $90^\circ$ can effectively avoid this problem, but it will introduce EoE problems.
    
\subsection{Densely Coded Label}
The introduction of excessive amount of parameters and the unfriendliness to small aspect ratio objects seriously hurts the applicability of classification based rotation detectors. In this section, we will solve the above problems from the two perspectives of encoding form and loss function weight.

Binary Coded Label (BCL) \cite{heath1972origins} and Gray Coded Label (GCL) \cite{frank1953pulse, yang2020on} are two Densely Coded Label (DCL) methods commonly used in the field of electronic communication. Their advantage is that they can represent a larger range of values with less coding length. Thus, they can effectively solve the problem of excessively long coding length in CSL and One-Hot based methods. The prediction layer thickness of BCL and GCL based methods are calculated as follows:
\begin{equation}
    \begin{aligned}
    \underbrace{\text{Th}_{bcl} = \text{Th}_{gcl}}_{\text{TH}_{dcl}} = & A \times \lceil \log_{2}({AR / \omega}) \rceil
    \label{eq:length}
    \end{aligned}
\end{equation}

Under the same setting of $A=21$, $AR=180$, $w=1$, the coding length of DCL is only $168$. According to Table \ref{table:flops_param}, GFlops and Param only increase by \textbf{3.24\%} and \textbf{0.92\%}. The training time is almost the same as the regression-based method. The performance of DCL-based method is not drop but rises, and the specific performance comparison can quickly refer to Figure \ref{fig:P0016_csl}-\ref{fig:P0016_gcl} and Table \ref{table:ablation_method}.

\begin{table}[tb!]
	\centering
	\resizebox{0.4\textwidth}{!}{
		\begin{tabular}{ccccccccccc}
			\toprule
			Decimal Number & 0 & 1 & 2 & 3 & 4 & 5 & 6 & 7\\
			\midrule
			
			Binary Coded Label & 000 & 001 & 010 & 011 & 100 & 101 & 110 & 111 \\
			Gray Coded Label & 000 & 001 & 011 & 010 & 110 & 111 & 101 & 100 \\

			\bottomrule
			
	\end{tabular}}
	\caption{The three-digit binary code and gray code corresponding to the decimal number.}
	\label{table:coding}
\end{table}

Algorithm \ref{alg:bcl}-\ref{alg:gcl} describe the pseudo codes that generate all $n$-bit gray and binary coded labels. BCL processes the angle by binarization to obtain a string of codes represented by multiple `0' and `1'. Although the coding length is greatly reduced, there may be huge changes in the coding results between adjacent values, that is, there is no classification tolerance mentioned in the CSL. For example, the three-bit binary coding results of the values `3' and `4' are `011' and `100', respectively. It can be seen that all three positions have changed, resulting in a very large difference in the loss value of the two angle predictions. GCL can solve this problem \cite{yang2020on}. In the encoding of a group of numbers, if any two adjacent codes differ only by one binary number, then this kind of encoding is called Gray Code. Due to only one digit is different between the maximum number and the minimum number, it is also called Cyclic Code. The coding results of `3' and `4' in the GCL method are `010' and `110'. Table \ref{table:coding} compares the coding results of BCL and GCL. The shortcomings of GCL are also obvious. Although the encoding forms between adjacent angles are not much different, which makes GCL also have a certain classification tolerance, the encoding differences of angles with large differences are not very significant, such as `1 (001)' and `6 (101)'. In summary, these two methods are agnostic or partially agnostic to the angle distance.

\begin{algorithm}[tb!]
  \caption{Binary Coded Label (BCL)}
  \label{alg:bcl}
  \begin{algorithmic}
  \STATE \textbf{Input:} angle range $AR$, discretization granularity $\omega$.
  \STATE \textbf{Output:} A list L containing all binary coded labels;
  \STATE $L = [ ]$, $n=\lceil \log_{2}({AR / \omega}) \rceil$ \# coding length;
  \FOR{i in $AR$}
  \STATE $bcl = Bin(i, n)$ \# generate $n$-bit binary code;
  \STATE $L.append(bcl)$;
  \ENDFOR
  \STATE \textbf{Return} $L$;
  
  \end{algorithmic}
\end{algorithm}

\begin{algorithm}[tb!]
  \caption{Gray Coded Label (GCL)}
  \label{alg:gcl}
  \begin{algorithmic}
  \STATE \textbf{Input:} coding length $n=\lceil \log_{2}({AR / \omega})\rceil$ ($AR$ means angle range, $\omega$ represents angle discretization granularity).
  \STATE \textbf{Output:} A list L containing all gray coded labels.
  \IF{$n=1$}
  \STATE \textbf{Return} $['0', '1']$;
  \ELSE
  \STATE $L1 = \text{GCL}(n-1)$ \# recursive call;
  \STATE $L2, L3, L4 = L1.reverse(), [ ], [ ]$;
  \FOR{s1 in L1} 
  \STATE $L3.append('0' + s1)$;
  \ENDFOR
  \FOR{s2 in L2} 
  \STATE $L3.append('1' + s2)$;
  \ENDFOR
  \STATE $L = concat(L3, L4)$;
  \ENDIF
  \STATE \textbf{Return} $L$;
  \end{algorithmic}
\end{algorithm}

In the DCL-based method, only the number of categories is a power of 2 to ensure that each coding corresponds to a valid angle. For example, if the 180 degree range is divided into $2^8=256$ categories, then the range of each division interval is $\omega=180/256=0.703125^\circ$. According to the $Max(error) = \omega / 2$ and $E(error)=\omega / 4$ proposed in work \cite{yang2020arbitrary}, the maximum and expected accuracy error are only $0.3515625^\circ$ and $0.17578125^\circ$, respectively whose influence on final detecton accuracy can be negligible. However, the above condition is not necessary. We find that even with some redundant invalid codes, there is no significant drop in final performance. Equation \ref{eq:encode_decode_dcl} specifies the encoding and decoding process of DCL (take BCL as an example):
\begin{equation}
	\begin{aligned}
	 \textbf{Encode:} & \ \text{Bin}( -\text{Round}((\theta_{gt}-90)/\omega)) \\
	 \textbf{Decode:} & \ 90 - \omega \text{Int}(\text{Round}(\text{Sigmoid}(logits)))
	\label{eq:encode_decode_dcl}
	\end{aligned}
\end{equation}

Figure \ref{fig:dcl} gives an example which also takes BCL as an embodiment. In the decoding process, the threshold for converting the predicted logits into binary coding is 0.5.

\subsection{Angle Distance and Aspect Ratio Sensitive Weighting}
To make the model sensitive to the distance of the angle, we calculate the decimal difference between the predicted angle and the angle label as an angle distance aware weight. The specific formula is designated as follows:
\begin{equation}
	\begin{aligned}
	  W(\Delta\theta) = & \log(|\Delta\theta|+1)= \log(|\theta_{gt}-\theta_{pred}|+1) \\
	  \theta_{pred} = & \text{Decode}_{dcl}(logits)
	\label{eq:ADW}
	\end{aligned}
\end{equation}
where $\Delta\theta$ denotes the decimal difference between the predicted angle ($\theta_{pred}$) and the angle label ($\theta_{gt}$). $logits$ represents the prediction vector of the angle.

However, the above-mentioned angle distance aware weight reintroduces the PoA problem. Take $\theta_{gt}=-90, \theta_{pred}=89$ as an example, although the angle of the two bounding boxes are very close, a very large weight is calculated. Therefore, we consider adding a periodic trigonometric function to solve this problem. As discussed in Section \ref{sec:scl}, the square-like object is not suitable to be defined by the long-side definition method. We propose an Angle Distance and Aspect Ratio Sensitive Weighting as ADARSW, as shown in Equation \ref{eq:ADARSW}:
\begin{equation}
	\begin{aligned}
     W_{ADARSW}(\Delta\theta) = & |\sin(\alpha(\Delta\theta))| = |\sin(\alpha(\theta_{gt}-\theta_{pred}))| \\
     \alpha = &
	\left\{ \begin{array}{rcl}
	1, & (h_{gt}/w_{gt})>r \\ 2, & otherwise
	\end{array}\right.
	\label{eq:ADARSW}
	\end{aligned}
\end{equation}
where $h_{gt}$ and $w_{gt}$ are the long and short sides of ground truth. $r$ is the aspect ratio threshold, the default value is 1.5. 

When the object has a certain aspect ratio, the period of $|\sin(\alpha(\theta_{gt}-\theta_{pred}))|$ is set to $180^\circ$ ($\alpha=1$), and when the object
is square-like, the period becomes $90^\circ$ ($\alpha=2$). Thus, the model can solve the PoA and can flexibly adjust the training strategy for different aspect ratio objects.
The DCL-based angle classification loss is as follows:
\begin{equation}
	\begin{aligned}
	L_{dcl}(\theta_{gt}, logits) = & \text{FL}(\text{Encode}_{dcl}(\theta_{gt}),logits) \\
	& \times W_{ADARSW}(\Delta\theta)
	\label{eq:dcl_loss}
	\end{aligned}
\end{equation}
where FL indicates focal loss \cite{lin2017focal}.

\subsection{Loss Function}
For RetinaNet-based rotation detection, we use five parameters ($x,y,h, w,\theta$) to represent arbitrary-oriented rectangle. Ranging in $[-\pi/2,\pi/2)$, the $\theta$ is determined by the long side ($h$) of the rectangle and x-axis. For DCL based method, it calls for an additional angle classification prediction layer. The other four parameters are predicted by regression, the regression formula is as follows:
\begin{equation}
	\begin{aligned}
	t_{x}&=(x-x_{a})/w_{a}, t_{y}=(y-y_{a})/h_{a} \\
	t_{w}&=\log(w/w_{a}), t_{h}=\log(h/h_{a})\\
	t_{x}^{'}&=(x_{}^{'}-x_{a})/w_{a}, t_{y}^{'}=(y_{}^{'}-y_{a})/h_{a} \\
	t_{w}^{'}&=\log(w_{}^{'}/w_{a}), t_{h}^{'}=\log(h_{}^{'}/h_{a})
	\label{eq:regression}
	\end{aligned}
\end{equation}
where $x,y,h,w$ denote the box's center coordinates, height and width respectively. Variables $x, x_{a}, x^{'}$ are for the ground-truth box, anchor box, and predicted box, respectively (likewise for $y,w,h$).

The multi-task loss is used which is defined as follows:
\begin{equation}
	\begin{aligned}
	L = & \frac{\lambda_{1}}{N}\sum_{n=1}^{N}obj_{n} \sum_{j\in\{x,y,h,w\}}L_{reg}(v_{nj}^{'},v_{nj}) \\	
	& + \frac{\lambda_{2}}{N}\sum_{n=1}^{N}obj_{n} L_{dcl}(\theta_{gt}, logits) + \frac{\lambda_{3}}{N}\sum_{n=1}^{N}L_{cls}(p_{n},t_{n}) 
	\label{eq:multitask_loss}
	\end{aligned}
\end{equation}
where $N$ indicates the number of anchors, $obj_{n}$ is a binary value ($obj_{n}=1$ for foreground and $obj_{n}=0$ for background, no regression for background). $v_{*j}^{'}$ denotes the predicted offset vectors, $v_{*j}$ is the targets vector of ground truth. $t_{n}$ represents the label of object, $p_{n}$ is the probability distribution of various classes calculated by sigmoid function. The hyper-parameter $\lambda_{1}$, $\lambda_{2}$, $\lambda_{3}$ control the trade-off and are set to $\{1,0.5,0.1\}$ by default. The classification loss $L_{cls}$ is focal loss \cite{lin2017focal}. The regression loss $L_{reg}$ is smooth L1 loss as used in \cite{girshick2015fast}.

\section{Experiments}\label{sec:experiments}
We use Tensorflow \cite{abadi2016tensorflow} to implement the proposed methods on a server with GeForce RTX 2080 Ti and 11G memory. The experiments in this article are initialized by ResNet50 \cite{he2016deep} by default unless otherwise specified. We perform experiments on both aerial benchmarks and scene text benchmarks to verify the generality of our techniques. Weight decay and momentum are set 0.0001 and 0.9, respectively. We employ MomentumOptimizer over 4 GPUs with a total of 4 images per minibatch (1 images per GPU).
\subsection{Datasets and Protocls}
DOTA \cite{xia2018dota} is comprised of 2,806 large aerial images from different sensors and platforms. Objects in DOTA exhibit a wide variety of scales, orientations, and shapes. These images are then annotated by experts using 15 object categories. 
The short names for categories are defined as (abbreviation-full name): PL-Plane, BD-Baseball diamond, BR-Bridge, GTF-Ground field track, SV-Small vehicle, LV-Large vehicle, SH-Ship, TC-Tennis court, BC-Basketball court, ST-Storage tank, SBF-Soccer-ball field, RA-Roundabout, HA-Harbor, SP-Swimming pool, and HC-Helicopter. 
The fully annotated DOTA benchmark contains 188,282 instances, each of which is labeled by an arbitrary quadrilateral. 
Half of the original images are randomly selected as the training set, 1/6 as the validation set, and 1/3 as the testing set. We divide the images into $ 600 \times 600 $ subimages with an overlap of 150 pixels and scale it to $ 800 \times 800 $. With all these processes, we obtain about 20,000 training and 7,000 validation patches.

UCAS-AOD \cite{zhu2015orientation} contains 1,510 aerial images of approximately $ 659 \times 1,280 $ pixels, with two categories of 14,596 instances in total. In line with~\cite{azimi2018towards,xia2018dota}, we randomly select 1,110 for training and 400 for testing. 

ICDAR2015 \cite{karatzas2015icdar} is the Challenge 4 of ICDAR 2015 Robust Reading Competition, which is commonly used for oriented scene text detection and spotting. This dataset includes 1,000 training images and 500 testing images.

ICDAR 2017 MLT \cite{nayef2017icdar2017} is a multi-lingual text dataset, which includes 7,200 training images, 1,800 validation images and 9,000 testing images. The dataset is composed of complete scene images in 9 languages, and text regions in this dataset can be in arbitrary orientations, being more diverse and challenging.

HRSC2016 \cite{liu2017high} contains images from two scenarios including ships on sea and ships close inshore. All images are collected from six famous harbors. The training, validation and test set include 436, 181 and 444 images, respectively.

\begin{figure}[!tb]
		\centering
		\subfigure[$\omega=180/4$]{
			\begin{minipage}[t]{0.47\linewidth}
				\centering
				\includegraphics[width=0.99\linewidth]{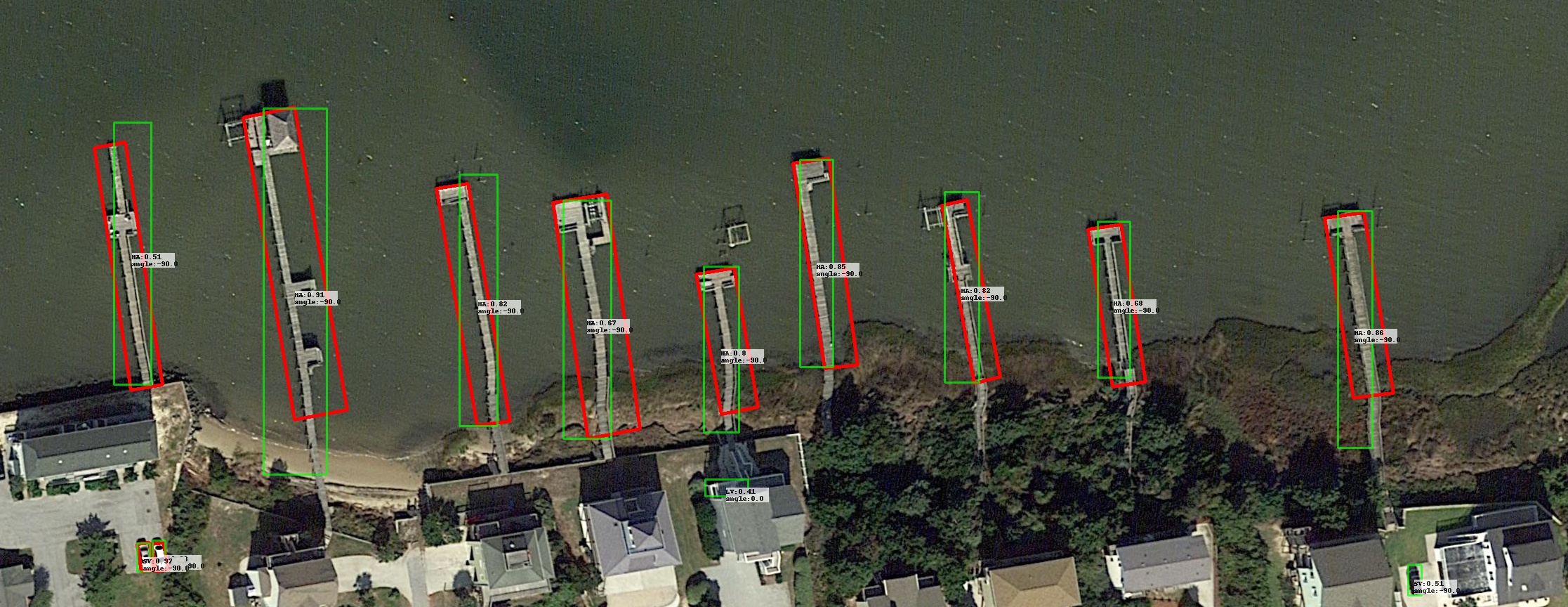}
			\end{minipage}%
			\label{fig:180/4}
		}
		\subfigure[$\omega=180/32$]{
			\begin{minipage}[t]{0.47\linewidth}
				\centering
				\includegraphics[width=0.99\linewidth]{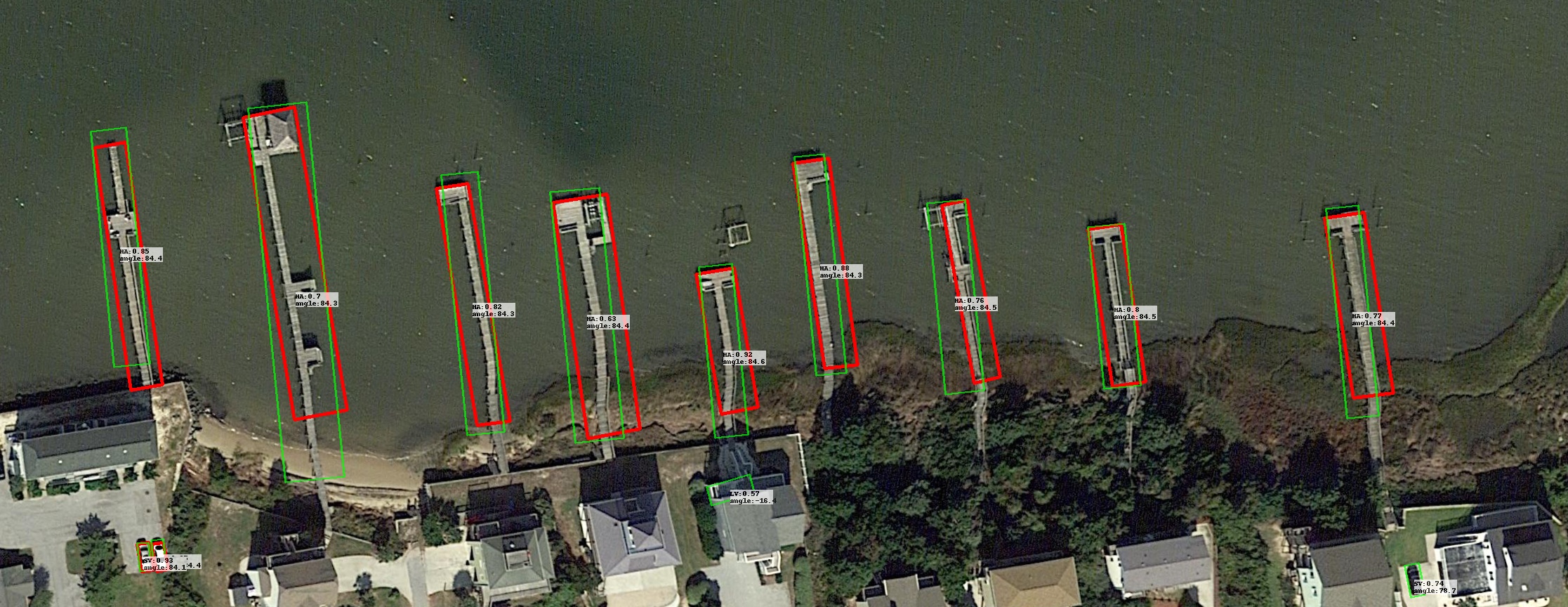}
			\end{minipage}%
			\label{fig:180/32}
		}\\
		\subfigure[$\omega=180/128$]{
			\begin{minipage}[t]{0.47\linewidth}
				\centering
				\includegraphics[width=0.99\linewidth]{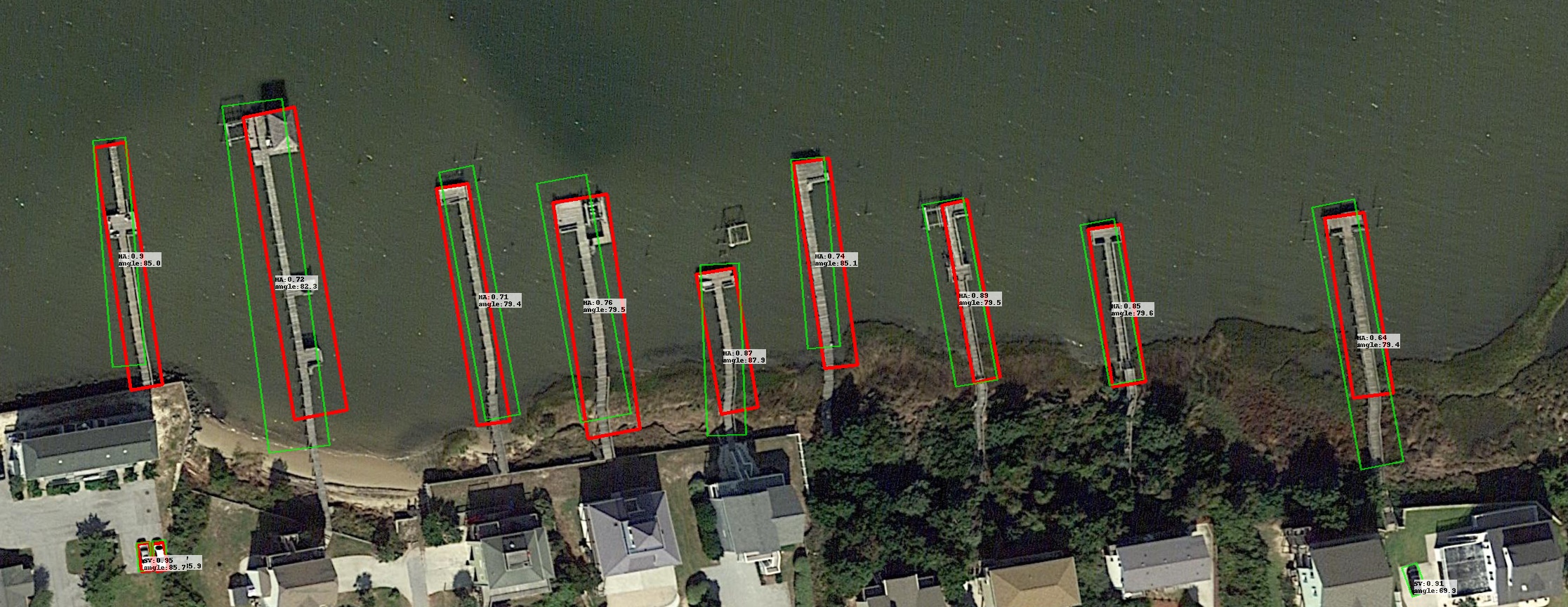}
			\end{minipage}
			\label{fig:180/128}
		}
		\subfigure[$\omega=180/256$]{
			\begin{minipage}[t]{0.47\linewidth}
				\centering
				\includegraphics[width=0.99\linewidth]{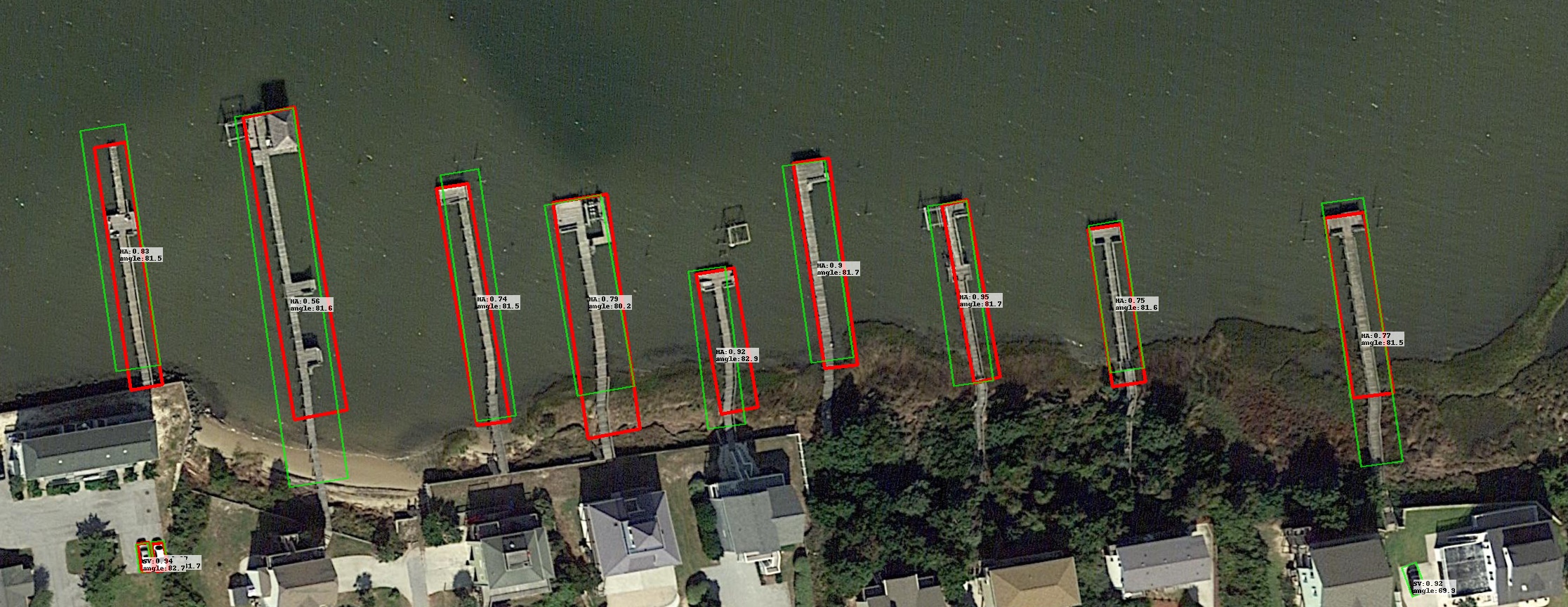}
			\end{minipage}%
			\label{fig:180/256}
		}
		\centering
		\caption{Visualization of detection results under different angle discretization granularity $\omega$. The red and green bounding box indicate ground truth and prediction.}
		\label{fig:omega_vis}
\end{figure}

\begin{table}[!tb]
	\centering
	\resizebox{0.48\textwidth}{!}{
		\begin{tabular}{c|ccccc|cc}
			\toprule
			Method & BR & SV & LV & SH & HA & 5-mAP$_{50}$ & mAP$_{50}$ \\
			\hline
			\multirow{1}{*}{\shortstack{RetinaNet-Reg}} & 38.31 & 60.48 & 49.77 & 68.29 & 51.28 & 53.63 & 64.17 \\
			\multirow{1}{*}{\shortstack{RetinaNet-CSL}} & 40.55 & 66.77 & 51.50 & 73.60 & 53.76 & 57.24 (+3.61) & 65.69 (+1.52) \\
            \multirow{1}{*}{\shortstack{RetinaNet-BCL}} & 41.58 & 67.98 & \textbf{57.34} & \textbf{74.66} & 54.28 & \textbf{59.17 (+5.54)} & \textbf{66.53 (+2.36)} \\
            \multirow{1}{*}{\shortstack{RetinaNet-GCL}} & \textbf{42.55} & \textbf{68.38} & 56.40 & 73.53 & \textbf{54.36} & 59.04 (+5.41) & 66.27 (2.10) \\
    
			\bottomrule
	\end{tabular}}
	\caption{Ablation study of four orientation detectors on DOTA test dataset. 5-mAP$_{50}$ means the performance of the five categories listed. The number in parentheses indicates the performance gain compared to the RetinaNet-Reg.}
	\label{table:ablation_method}
\end{table}

\begin{table}[!tb]
	\centering
	\resizebox{0.48\textwidth}{!}{
		\begin{tabular}{c|c|ccccc|cccc}
			\toprule
			Method & $\omega$ & BR & SV & LV & SH & HA & 5-mAP$_{50}$ & mAP$_{50}$ & mAP$_{75}$ & mAP$_{50:95}$ \\
			\hline
			\multirow{1}{*}{\shortstack{Reg}} & - & 34.52 & 51.42 & 50.32 & 73.37 & 55.93 & 53.12 & 62.21 & 26.07 & 31.49\\
			\hline
			\multirow{1}{*}{\shortstack{CSL}} & 180/180 & 35.94 & 53.42 & 61.06 & 81.81 & 62.14 & 58.87 & 64.40 & 32.58 & 35.04\\
			\hline
            \multirow{8}{*}{\shortstack{BCL}}
            & 180/4 & 30.74 & 40.54 & 50.98 & 72.07 & 59.54 & 50.77 & 62.38 & 24.88 & 31.01 \\
            & 180/8 & 36.65 & 52.58 & 60.46 & 82.24 & 61.60 & 58.71 & \textbf{66.17} & 33.14 & 35.77 \\
            & 180/32 & \textbf{39.83} & 54.41 & 60.62 & 80.81 & 60.32 & \textbf{59.20} & 65.93 & \textbf{35.66} & \textbf{36.71} \\
            & 180/64 & 38.22 & \textbf{54.70} & 60.16 & 80.75 & 60.11 & 58.79 & 65.00 & 34.31 & 36.00 \\
            & 180/128 & 36.76 & 53.73 & \textbf{61.35} & \textbf{82.52} & 58.42 & 58.56 & 65.14 & 34.28 & 35.69 \\
            & 180/180 & 37.42 & 53.72 & 58.70 & 80.73 & \textbf{63.31} & 58.78 & 65.83 & 33.94 & 36.35 \\
            & 180/256 & 37.66 & 53.83 & 60.66 & 80.43 & 60.74 & 58.66 & 64.97 & 33.52 & 35.21 \\
            & 180/512 & 37.93 & 53.85 & 58.52 & 80.04 & 60.87 & 58.24 & 64.88 & 33.09 & 34.99 \\
            \hline
            \multirow{7}{*}{\shortstack{GCL}}
            & 180/4 & 30.90 & 41.20 & 48.30 & 72.93 & 60.16 & 50.70 & 62.98 & 23.83 & 30.81 \\
            & 180/8 & 36.88 & 51.10 & 59.81 & 82.40 & 61.57 & 58.35 & 65.23 & 33.92 & 35.29 \\
            & 180/32 & 38.04 & \textbf{54.77} & 60.88 & \textbf{82.75} & 61.24 & \textbf{59.54} & 65.11 & \textbf{34.67} & 36.15 \\
            & 180/64 & \textbf{38.05} & 54.36 & 60.59 & 81.84 & 60.39 & 59.05 & 64.78 & 33.23 & 35.67 \\
            & 180/128 & 37.74 & 54.36 & 59.43 & 81.15 & 60.51 & 58.64 & \textbf{66.13} & 33.65 & \textbf{36.34} \\
            & 180/256 & 35.81 & 53.78 & 58.35 & 81.45 & 59.84 & 57.85 & 64.87 & 33.77 & 35.97 \\
            & 180/512 & 37.99 & 54.23 & \textbf{61.61} & 80.84 & \textbf{62.13} & 59.36 & 64.34 & 34.08 & 35.92 \\

			\bottomrule
	\end{tabular}}
	\caption{Comparison of detection results under different angle discretization granularities denoted by $\omega$.}
	\label{table:ablation_omega}
\end{table}

\begin{table}[!tb]
	\centering
	\resizebox{0.48\textwidth}{!}{
		\begin{tabular}{cccccccccccc|cc}
			\toprule
			Method & ADARSW & PL & BD & GTF & TC & BC & ST & SBF & RA & SP & HC & 10-mAP$_{50}$ & mAP$_{50}$ \\
			\hline
			\multirow{2}{*}{BCL}
			 & & 88.63 & 71.62 & 65.18 & 90.70 & 76.32 & 78.47 & 52.26 & 60.25 & 66.61 & 49.15 & 69.92 & 66.53\\
			 & $\checkmark$ & 88.92 & 72.11 & 66.32 & 90.79 & 79.86 & 79.03 & 54.11 & 63.18 & 67.86 & 60.04 & \textbf{72.22} & \textbf{67.39}\\
			 \hline
			 \multirow{2}{*}{GCL}
			 & & 88.52 & 73.58 & 64.38 & 90.80 & 77.66 & 76.38 & 50.84 & 59.46 & 65.83 & 48.42 & 69.59 & 66.27\\
			 & $\checkmark$ & 88.96 & 75.20 & 65.24 & 90.78 & 79.13 & 77.95 & 55.60 & 61.90 & 66.18 & 56.27 & \textbf{71.72} & \textbf{67.02}\\ 

			\bottomrule
	\end{tabular}}
	\caption{The verification of ADARSW on small aspect ratio objects in the DOTA dataset.}
	\label{table:ablation_ad2pw}
\end{table}

\begin{table*}[tb!]
	\centering
	\resizebox{0.9\textwidth}{!}{
			\begin{tabular}{c|ccc|cccc|ccc}
				\toprule
				\multirow{2}{*}{Method}&\multicolumn{3}{c}{ICDAR2015}&\multicolumn{4}{c}{UCAS-AOD}&\multicolumn{3}{c}{MLT} \\
				\cmidrule(lr){2-4} \cmidrule(lr){5-8} \cmidrule(lr){9-11}
				& Recall & Precision & Hmean & car(07/12) & plane(07/12) & mAP$_{50}$ (07) & mAP$_{50}$ (12) & Recall & Precision & Hmean\\
				\hline
				RetinaNet-Reg & 81.49 & 83.29 & 82.38 & 87.28/90.79 & 90.42/97.52 &	88.85 & 94.16 & 55.70 &	\textbf{75.24} & 64.01 \\
				RetinaNet-CSL & 80.50 & \textbf{87.40} & \textbf{83.81 (+1.43)} & 88.09/\textbf{92.93} & 90.38/97.22 & 89.23 (+0.38) & 95.07 (+0.91) & 58.32 & 73.62 & 65.08 (+1.07) \\
				RetinaNet-BCL & \textbf{81.61} & 84.79 & 83.17 (+0.79) & \textbf{88.15}/92.35 & \textbf{90.57}/\textbf{97.86} & \textbf{89.36 (+0.51)} & \textbf{95.10 (+0.94)} & \textbf{58.91} & 73.14 & \textbf{65.26 (+1.25)} \\
				\bottomrule
			\end{tabular}}
			\caption{Comparison between classification-based and regression-based methods on more datasets. 2007 or 2012 in bracket means using the 2007 or 2012 evaluation metric. ResNet101, data augmentation, multi-scale training and testing are used.}
			\label{table:other_dataset_comparison}
\end{table*}

\begin{table*}[tb!]
		\centering
		\resizebox{0.98\textwidth}{!}{
			\begin{tabular}{l|lcccccccccccccccccc}
				\toprule
				& Method & Backbone & MS &  PL &  BD &  BR &  GTF &  SV &  LV &  SH &  TC &  BC &  ST &  SBF &  RA &  HA &  SP &  HC &  mAP$_{50}$\\
				\midrule
				\multirow{22}{*}{\rotatebox{270}{\shortstack{Two-stage methods}}} 
				&FR-O \cite{xia2018dota} & ResNet101 & & 79.09 & 69.12 & 17.17 & 63.49 & 34.20 & 37.16 & 36.20 & 89.19 & 69.60 & 58.96 & 49.4 & 52.52 & 46.69 & 44.80 & 46.30 & 52.93 \\
				&R-DFPN \cite{yang2018automatic} & ResNet101 & & 80.92 & 65.82 & 33.77 & 58.94 & 55.77 & 50.94 & 54.78 & 90.33 & 66.34 & 68.66 & 48.73 & 51.76 & 55.10 & 51.32 & 35.88 & 57.94 \\
				&R$^2$CNN \cite{jiang2017r2cnn} & ResNet101 & & 80.94 & 65.67 & 35.34 & 67.44 & 59.92 & 50.91 & 55.81 & 90.67 & 66.92 & 72.39 & 55.06 & 52.23 & 55.14 & 53.35 & 48.22 & 60.67 \\
				&RRPN \cite{ma2018arbitrary} & ResNet101 & & 88.52 & 71.20 & 31.66 & 59.30 & 51.85 & 56.19 & 57.25 & 90.81 & 72.84 & 67.38 & 56.69 & 52.84 & 53.08 & 51.94 & 53.58 & 61.01 \\
				&ICN \cite{azimi2018towards} & ResNet101 & $\checkmark$ & 81.40 & 74.30 & 47.70 & 70.30 & 64.90 & 67.80 & 70.00 & 90.80 & 79.10 & 78.20 & 53.60 & 62.90 & 67.00 & 64.20 & 50.20 & 68.20 \\
				&RADet \cite{li2020radet} & ResNeXt101 \cite{xie2017aggregated} & & 79.45 & 76.99 & 48.05 & 65.83 & 65.46 & 74.40 & 68.86 & 89.70 & 78.14 & 74.97 & 49.92 & 64.63 & 66.14 & 71.58 & 62.16 & 69.09 \\
				&RoI-Transformer \cite{ding2018learning} & ResNet101 & $\checkmark$ & 88.64 & 78.52 & 43.44 & 75.92 & 68.81 & 73.68 & 83.59 & 90.74 & 77.27 & 81.46 & 58.39 & 53.54 & 62.83 & 58.93 & 47.67 & 69.56 \\
				&CAD-Net \cite{zhang2019cad} & ResNet101 & & 87.8 & 82.4 & 49.4 & 73.5 & 71.1 & 63.5 & 76.7 & 90.9 & 79.2 & 73.3 & 48.4 & 60.9 & 62.0 & 67.0 & 62.2 & 69.9 \\
				&AOOD \cite{zou2020arbitrary} & DPN \cite{chen2017dual} & $\checkmark$ & 89.99 & 81.25 & 44.50 & 73.20 & 68.90 & 60.33 & 66.86 & 90.89 & 80.99 & 86.23 & 64.98 & 63.88 & 65.24 & 68.36 & 62.13 & 71.18 \\
				&Cascade-FF \cite{hou2020cascade} & ResNet152 & & 89.9 & 80.4 & 51.7 & \textbf{77.4} & 68.2 & 75.2 & 75.6 & 90.8 & 78.8 & 84.4 & 62.3 & 64.6 & 57.7 & 69.4 & 50.1 & 71.8	\\
				&SCRDet \cite{yang2019scrdet} & ResNet101 & $\checkmark$ & 89.98 & 80.65 & 52.09 & 68.36 & 68.36 & 60.32 & 72.41 & 90.85 & \textbf{87.94} & 86.86 & 65.02 & 66.68 & 66.25 & 68.24 & 65.21 & 72.61\\
				&SARD \cite{wang2019sard} & ResNet101 & & 89.93 & 84.11 & 54.19 & 72.04 & 68.41 & 61.18 & 66.00 & 90.82 & 87.79 & 86.59 & 65.65 & 64.04 & 66.68 & 68.84 & 68.03 & 72.95 \\
				&GLS-Net \cite{li2020object} & ResNet101 & & 88.65 & 77.40 & 51.20 & 71.03 & 73.30 & 72.16 & 84.68 & 90.87 & 80.43 & 85.38 & 58.33 & 62.27 & 67.58 & 70.69 & 60.42 & 72.96 \\
				&FADet \cite{li2019feature} & ResNet101 & $\checkmark$ & 90.21 & 79.58 & 45.49 & 76.41 & 73.18 & 68.27 & 79.56 & 90.83 & 83.40 & 84.68 & 53.40 & 65.42 & 74.17 & 69.69 & 64.86 & 73.28\\
				&MFIAR-Net \cite{yang2020multi} & ResNet152 & $\checkmark$ & 89.62 & 84.03 & 52.41 & 70.30 & 70.13 & 67.64 & 77.81 & 90.85 & 85.40 & 86.22 & 63.21 & 64.14 & 68.31 & 70.21 & 62.11 & 73.49 \\
				&Gliding Vertex \cite{xu2020gliding} & ResNet101 & & 89.64 & 85.00 & 52.26 & 77.34 & 73.01 & 73.14 & 86.82 & 90.74 & 79.02 & 86.81 & 59.55 & \textbf{70.91} & 72.94 & 70.86 & 57.32 & 75.02 \\
				&Mask OBB \cite{wang2019mask} & ResNeXt101 & $\checkmark$ & 89.56 & 85.95 & 54.21 & 72.90 & 76.52 & 74.16 & 85.63 & 89.85 & 83.81 & 86.48 & 54.89 & 69.64 & 73.94 & 69.06 & 63.32 & 75.33 \\
				&FFA \cite{fu2020rotation} & ResNet101 & $\checkmark$ & 90.1 & 82.7 & 54.2 & 75.2 & 71.0 & 79.9 & 83.5 & 90.7 & 83.9 & 84.6 & 61.2 & 68.0 & 70.7 & 76.0 & 63.7 & 75.7 \\
				&APE \cite{zhu2020adaptive} & ResNeXt101 & & 89.96 & 83.62 & 53.42 & 76.03 & 74.01 & 77.16 & 79.45 & 90.83 & 87.15 & 84.51 & 67.72 & 60.33 & \textbf{74.61} & 71.84 & 65.55 & 75.75 \\
				&CenterMap OBB \cite{wang2020learning} & ResNet101 & $\checkmark$ & 89.83 & 84.41 & 54.60 & 70.25 & 77.66 & 78.32 & 87.19 & 90.66 & 84.89 & 85.27 & 56.46 & 69.23 & 74.13 & 71.56 & 66.06 & 76.03\\
				&FPN-CSL \cite{yang2020arbitrary} & ResNet152 & $\checkmark$ & \textbf{90.25} & 85.53 & 54.64 & 75.31 & 70.44 & 73.51 & 77.62 & 90.84 & 86.15 & 86.69 & 69.60 & 68.04 & 73.83 & 71.10 & 68.93 & 76.17\\
				&SCRDet++ \cite{yang2020scrdet++} & ResNet101 & $\checkmark$ & 90.05 & 84.39 & \textbf{55.44} & 73.99 & 77.54 & 71.11 & 86.05 & 90.67 & 87.32 & \textbf{87.08} & \textbf{69.62} & 68.90 & 73.74 & 71.29 & 65.08 & 76.81\\
				\hline
				\multirow{13}{*}{\rotatebox{270}{\shortstack{Single-stage methods}}} 
				&IENet \cite{lin2019ienet} & ResNet101 & $\checkmark$ & 80.20 & 64.54 & 39.82 & 32.07 & 49.71 & 65.01 & 52.58 & 81.45 & 44.66 & 78.51 & 46.54 & 56.73 & 64.40 & 64.24 & 36.75 & 57.14 \\
				&TOSO \cite{feng2020toso} & ResNet101 & $\checkmark$ & 80.17 & 65.59 & 39.82 & 39.95 & 49.71 & 65.01 & 53.58 & 81.45 & 44.66 & 78.51 & 48.85 & 56.73 & 64.40 & 64.24 & 36.75 & 57.92\\
				&PIoU \cite{chen2020piou} & DLA-34 \cite{chen2020piou} & & 80.9 & 69.7 & 24.1 & 60.2 & 38.3 & 64.4 & 64.8 & 90.9 & 77.2 & 70.4 & 46.5 & 37.1 & 57.1 & 61.9 & 64.0 & 60.5 \\
				&P-RSDet \cite{zhou2020objects} & ResNet101 & $\checkmark$ & 89.02 & 73.65 & 47.33 & 72.03 & 70.58 & 73.71 & 72.76 & 90.82 & 80.12 & 81.32 & 59.45 & 57.87 & 60.79 & 65.21 & 52.59 & 69.82 \\
				&O$^2$-DNet \cite{wei2020oriented} & Hourglass104 \cite{newell2016stacked} & $\checkmark$ & 89.31 & 82.14 & 47.33 & 61.21 & 71.32 & 74.03 & 78.62 & 90.76 & 82.23 & 81.36 & 60.93 & 60.17 & 58.21 & 66.98 & 61.03 & 71.04 \\
				&BBAVectors \cite{yi2020oriented} & ResNet101 & $\checkmark$ & 88.35 & 79.96 & 50.69 & 62.18 & 78.43 & 78.98 & \textbf{87.94} & 90.85 & 83.58 & 84.35 & 54.13 & 60.24 & 65.22 & 64.28 & 55.70 & 72.32 \\
				&DRN \cite{pan2020dynamic} & Hourglass104 & $\checkmark$ & 89.71 & 82.34 & 47.22 & 64.10 & 76.22 & 74.43 & 85.84 & 90.57 & 86.18 & 84.89 & 57.65 & 61.93 & 69.30 & 69.63 & 58.48 & 73.23 \\
				&R$^3$Det \cite{yang2019r3det} & ResNet152 &  & 89.49 & 81.17 & 50.53 & 66.10 & 70.92 & 78.66 & 78.21 & 90.81 & 85.26 & 84.23 & 61.81 & 63.77 & 68.16 & 69.83 & 67.17 & 73.74 \\
				&RSDet \cite{qian2019learning} & ResNet152 & & 90.1 & 82.0 & 53.8 & 68.5 & 70.2 & 78.7 & 73.6 & \textbf{91.2} & 87.1 & 84.7 & 64.3 & 68.2 & 66.1 & 69.3 & 63.7 & 74.1 \\
				&PolarDet \cite{zhao2020polardet} & ResNet101 & $\checkmark$ & 89.65 & \textbf{87.07} & 48.14 & 70.97 & 78.53 & 80.34 & 87.45 & 90.76 & 85.63 & 86.87 & 61.64 & 70.32 & 71.92 & 73.09 & 67.15 & 76.64\\
				\cline{2-20}
				&RetinaNet-DCL (Ours) & ResNet152 & $\checkmark$ & 89.10 & 84.13 & 50.15 & 73.57 & 71.48 & 58.13 & 78.00 & 90.89 & 86.64 & 86.78 & 67.97 & 67.25 & 65.63 & \textbf{74.06} & 67.05 & 74.06 \\
				&R$^3$Det-DCL (Ours) & ResNet152 &  & 89.78 & 83.95 & 52.63 & 69.70 & 76.84 & 81.26 & 87.30 & 90.81 & 84.67 & 85.27 & 63.50 & 64.16 & 68.96 & 68.79 & 65.45 & 75.54\\
				&R$^3$Det-DCL (Ours) & ResNet101 & $\checkmark$ & 89.14 & 83.93 & 53.05 & 72.55 & 78.13 & 81.97 & 86.94 & 90.36 & 85.98 & 86.94 & 66.19 & 65.66 & 73.72 & 71.53 & 68.69 & 76.97\\
				&R$^3$Det-DCL (Ours) & ResNet152 & $\checkmark$ & 89.26 & 83.60 & 53.54 & 72.76 & \textbf{79.04} & \textbf{82.56} & 87.31 & 90.67 & 86.59 & 86.98 & 67.49 & 66.88 & 73.29 & 70.56 & \textbf{69.99} & \textbf{77.37} \\
				\bottomrule
		\end{tabular}}
		\caption{Detection accuracy on different objects (AP$_{50}$) and overall performance (mAP$_{50}$) evaluation on DOTA. MS indicates that multi-scale training or testing is used. }
		\label{table:DOTA_OBB}
\end{table*}

All the used datasets are trained by 20 epochs in total, and learning rate is reduced tenfold at 12 epochs and 16 epochs, respectively. The initial learning rates for RetinaNet is 5e-4. The number of image iterations per epoch for DOTA, UCAS-AOD, HRSC2016, ICDAR2015, and MLT are 27k, 5k, 10k, 10k, 10k and 10k respectively, and doubled if data augmentation and multi-scale training are used. 

\subsection{Ablation Study}\label{sec:ablation_study}
\textbf{Comparison of four object orientation detectors.}
Table \ref{table:ablation_method} compares the performance of a regression based detector: RetinaNet-Reg and three classification based detectors: RetinaNet-CSL, RetinaNet-BCL and RetinaNet-GCL. We mainly focus on comparing five categories with large aspect ratios and more boundary conditions. It can be clearly seen that classification based detector outperforms those based on regression, with about 1.5\%-2.3\% and 3.6\%-5.5\% gain in overall performance (mAP$_{50}$) and five categories performance (5-mAP$_{50}$). More importantly, the performance of the DCL-based detector is nearly three times faster than the CSL-based detector, and the performance can still be further improved by about 2\% and 0.8\% in 5-mAP$_{50}$ and mAP$_{50}$. Figure \ref{fig:bc_vis} shows the visual qualitative comparison of the four methods under boundary conditions. We can fully draw the conclusion that the orientation estimation based on classification is a boundary discontinuity free method.

\textbf{Angle discretization granularity.}
In general, the smaller $\omega$, the higher theoretical upper bound of the model's performance. However, the decrease of $\omega$ will lead to an increase in the number of angle categories, which poses a challenge to the angle classification performance of the model. Therefore, we need to explore the impact of $\omega$ on the detection performance under different IoU thresholds, and find a suitable range of $\omega$. In order to get the performance indicators under different IoU threshold, we conduct experiments on the DOTA validation set, and the number of image iterations per epoch is 20k. According to Table \ref{table:ablation_omega}, when the number of angle categories is between 32 and 128, the performance of the model reaches its peak. If the number of categories is too small, the theoretical accuracy loss is too large, resulting in a sharp drop in performance; if the number of categories is too large, the angle classification network of the model cannot be effectively processed and the performance will decrease slightly. Figure \ref{fig:omega_vis} shows the comparison of angle estimates under different $\omega$.

\textbf{Redundant invalid coding.}
To make each code have a corresponding different angle value, the number of categories must be a power of 2 in the DCL-based method. However, this is not required. When we only set 180 categories, about 76 codings are invalid, but BCL-based method can still achieve good performance, at 36.35\% as shown in Table \ref{table:ablation_omega}. We also artificially increase the length based on the theoretical shortest code length to increase the proportion of invalid codes, and the performance is only slightly reduced.

\textbf{Angle Distance and Aspect Ratio Sensitive Weighting.} 
We mainly focus on comparing ten categories with small aspect ratio in Table \ref{table:ablation_ad2pw} to verify the effectiveness of ADARSW.
Under the same environment and hyperparameters, we add ADARSW to the BCL and GCL based methods, which and increase by 2.3\% and 2.13\% in ten categories performance (10-mAP$_{50}$). The overall performance has also increased to 67.39\% and 67.02\%. After ADARSW is used, the model can predict the bounding box as shown in Figure \ref{fig:pred_180_BD}.

\textbf{Comparison on more datasets and detectors.}
Table~\ref{table:other_dataset_comparison} further verifies the performance advantage of DCL based methods than CSL and regression based method on more datasets, including the text dataset ICDAR2015, MLT, and another remote sensing dataset UCAS-AOD. It is worth noting that the comparison results are based on large backbone, data augmentation, and multi-scale training and testing. We can still draw the conclusions: classification is better than regression for orientation estimation; DCL outperforms CSL in most cases. In order to verify the portability of DCL, we also conduct experiments on R$^3$Det. As shown in Table \ref{table:DOTA_OBB}, DCL can still make R$^3$Det get 1.8\% improvement under the use of large backbone and data augmentation.

\begin{table}[tb!]
		\centering
		\resizebox{0.45\textwidth}{!}{
			\begin{tabular}{lccc}
				\toprule
				
				Method & Backbone & mAP (07) & mAP (12)\\
				
				\midrule
                R$^2$CNN \cite{jiang2017r2cnn} & ResNet101 & 73.07 & 79.73 \\
				RC1 \& RC2 \cite{liu2017high} & VGG16 & 75.7 & -- \\
				RRPN \cite{ma2018arbitrary} & ResNet101 & 79.08 & 85.64 \\
				R$^2$PN \cite{zhang2018toward}  & VGG16 & 79.6 & --  \\
				RetinaNet-H \cite{yang2019r3det} & ResNet101 & 82.89 & 89.27 \\
				RRD \cite{liao2018rotation} & VGG16  & 84.3 & --  \\
				RoI-Transformer \cite{ding2018learning} & ResNet101 & 86.20 & -- \\
				Gliding Vertex \cite{xu2020gliding} & ResNet101 & 88.20 & -- \\
				BBAVectors \cite{yi2020oriented} & ResNet101 & 88.6 & -- \\
				DRN \cite{pan2020dynamic} & Hourglass104 & -- & 92.70 \\
				CenterMap OBB \cite{wang2020learning} & ResNet50 & -- & 92.8 \\
				SBD \cite{liu2019omnidirectional} & ResNet50 & -- & 93.70 \\
				RetinaNet-R \cite{yang2019r3det} & ResNet101 & 89.18 & 95.21 \\
				R$^3$Det \cite{yang2019r3det} & ResNet101 & 89.26 & 96.01 \\
				\hline
                R$^3$Det-DCL (Ours) & ResNet101 & \textbf{89.46} & \textbf{96.41}\\
				\bottomrule
				
		\end{tabular}}
		\caption{Detection accuracy on HRSC2016. 07 (12) means using the 2007(2012) evaluation metric.}
		\label{table:HRSC2016}
\end{table}

\begin{figure}[!tb]
		\centering
		\subfigure[$\omega=180/4$]{
			\begin{minipage}[t]{0.47\linewidth}
				\centering
				\includegraphics[width=0.98\linewidth]{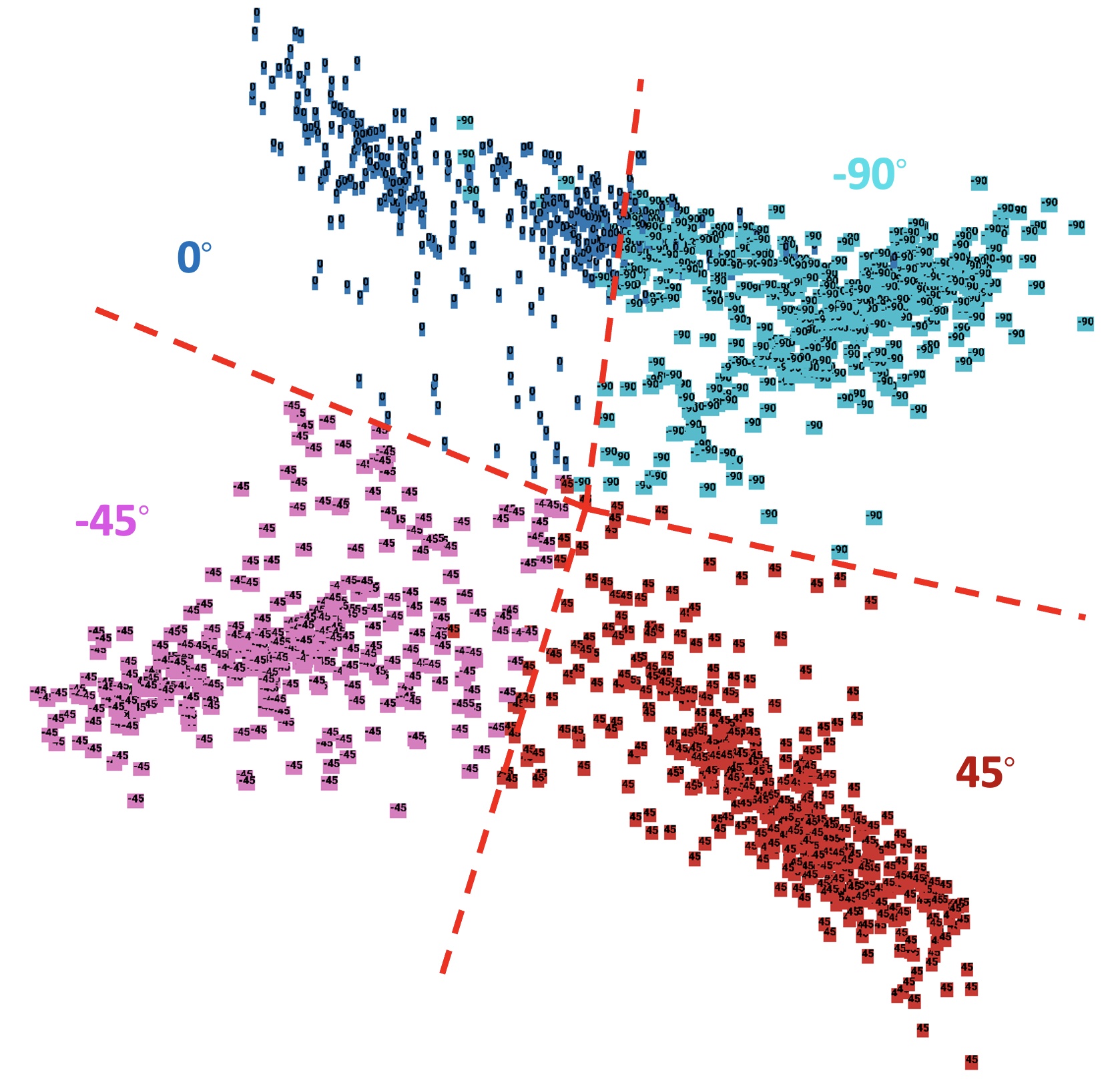}
			\end{minipage}%
			\label{fig:feature_vis_4}
		}
		\subfigure[$\omega=180/8$]{
			\begin{minipage}[t]{0.47\linewidth}
				\centering
				\includegraphics[width=0.98\linewidth]{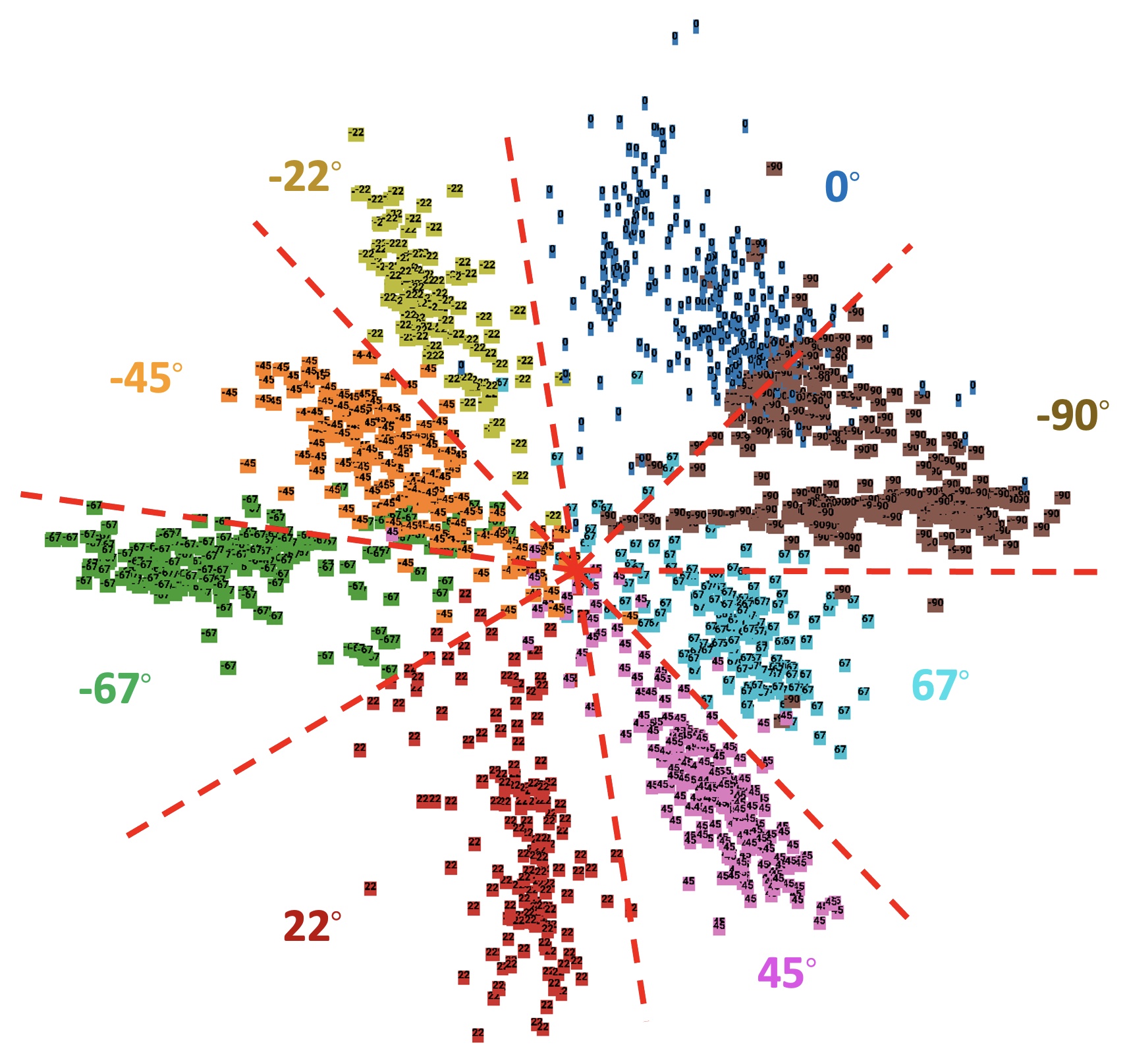}
			\end{minipage}%
			\label{fig:feature_vis_8}
		}
		\centering
		\caption{Angular feature visualization of the RetinaNet-DCL. The red dotted lines divide the different categories.}
		\label{fig:feature_vis}
\end{figure}

\textbf{Visual analysis of angular features.}
To further analyze the angle classification ability of the model, we use the principal component analysis (PCA) \cite{wold1987principal} to visualize each positive angle feature vector. We show the visualization results when the number of angle categories are 4 and 8, as shown in Figure \ref{fig:feature_vis}. This proves that it is feasible to use classification for orientation estimation, even if only the simplest cross-entropy loss function is used.

\subsection{Comparison with the State-of-the-Art}
We choose DOTA as the main comparison dataset due to the complexity of the aerial image and the large number of small, cluttered and rotated objects. As shown in Table \ref{table:DOTA_OBB}, through data augmentation, multi-scale training and testing commonly used by other advanced methods, RetinaNet-DCL-ResNet152 and R$^3$Det-DCL-ResNet152 can achieve competitive performance, about 74.06\% and 77.37\%, respectively.

The HRSC2016 contains lots of large aspect ratio ship instances with arbitrary orientation, which poses a huge challenge to the positioning accuracy of the detector. Experimental results show that our model achieves state-of-the-art performances, about 89.46\% (96.41\%).

\section{Conclusion}
This paper develops the line of research in classification based methodology for rotation detection in two folds: i) for the prediction layer, two Densely Coded Labels (DCL) techniques are devised by shortening the code length to achieve a more light-weighted prediction layer. They both accelerate the training speed of the recently proposed Sparsely Coded Label model in orientation classification based detectors notably. ii) We further propose the technique called Angle Distance and Aspect Ratio Sensitive Weighting (ADARSW), which further improves the performance by making DCL-based detector sensitive to angular distance and object's aspect ratio. Extensive experiments on different detectors and datasets show competitive performance regarding with both accuracy and efficiency.

{\small
\bibliographystyle{ieee_fullname}
\bibliography{egbib}
}

\end{document}